\documentclass[12pt]{article}
\usepackage{amsmath,amssymb,amsthm,thmtools,bm,enumerate,natbib,xr,float}
\usepackage{graphicx,psfrag,epsf}
\usepackage[dvipsnames]{xcolor}
\usepackage{url} 

\newcommand{\blind}{0}

\numberwithin{equation}{section}
\theoremstyle{plain}
\newtheorem{theorem}{Theorem}[section]

\newtheorem{corollary}{Corollary}[section]

\newtheorem{assumption}{}

\theoremstyle{remark}

\newcommand{\R}{\mathbb{R}}
\newcommand{\E}{\mathbb{E}}
\newcommand{\V}[1]{{\bm{\mathbf{\MakeLowercase{#1}}}}} 

\begin{document}

\def\spacingset#1{\renewcommand{\baselinestretch}%
{#1}\small\normalsize} \spacingset{1}


\if0\blind
{
  \title{\bf Statistical Inference for Online Decision-Making : In a Contextual Bandit Setting}
  \author{Haoyu Chen, Wenbin Lu and Rui Song\thanks{
  		Haoyu Chen is graduate student (E-mail: hchen36@ncsu.edu), Wenbin Lu is Professor (E-mail: lu@stat.ncsu.edu), and Rui Song is Professor (rsong@ncsu.edu), Department
  		of Statistics, North Carolina State University, Raleigh, NC 27695. The authors thank the editor, the AE and two referees for their helpful suggestions that significantly improved the quality of the paper. The research of Wenbin Lu is partially supported by Grant NCI P01 CA142538. The research of Rui Song is partially supported by Grant NSF-DMS-1555244 and Grant NCI P01 CA142538. }	
\date{\empty}
}
  \maketitle
} \fi

\if1\blind
{
  \bigskip
  \bigskip
  \bigskip
  \begin{center}
    {\LARGE\bf Statistical Inference for Online Decision-Making: In a Contextual Bandit Setting}
\end{center}
  \medskip
} \fi

\bigskip
\begin{abstract}
	Online decision-making problem requires us to make a sequence of decisions based on incremental information. Common solutions often need to learn a reward model of different actions given the contextual information and then maximize the long-term reward. It is meaningful to know if the posited model is reasonable and how the model performs in the asymptotic sense. We study this problem under the setup of the contextual bandit framework with a linear reward model. The $\varepsilon$-greedy policy is adopted to address the classic exploration-and-exploitation dilemma. Using the martingale central limit theorem, we show that the online ordinary least squares estimator of  model parameters is asymptotically normal. When the linear model is misspecified, we propose the online weighted least squares estimator using the inverse propensity score weighting and also establish its asymptotic normality. Based on the properties of the parameter estimators, we further show that the in-sample inverse propensity weighted value estimator is asymptotically normal. We illustrate our results using simulations and an application to a news article recommendation dataset from Yahoo!. 
\end{abstract}
		
\noindent%
\vfill

\newpage
\spacingset{1.5} 
\section{Introduction}
In this era of personalization, every decision-maker wants to exploit personal information to optimize the decisions. In scenarios from clinical trials \citep{bather1985allocation, armitage1985search}, to news article recommendation \citep{li2010contextual}, it is often the case that people will arrive sequentially, thus decision-makers will have to make a series of decisions with incremental information. A highly simplified model for such sequential decision-making with covariates is first formulated by \citet{woodroofe1979one}, and later named ``contextual bandit" by \citet{langford2008epoch}. The problem can be described as choosing one of the available actions at each time point when a person arrives, where each action will produce a random reward. The goal is to maximize the cumulative rewards over a long period with the help of each person's features. The task here is two-fold: we need to learn the relationship between features and optimal decisions and we need to learn it in an online fashion. If the whole problem is offline, many methods have been proposed to solve it \citep[e.g.][]{murphy2003optimal, zhao2012estimating, zhang2012robust, zhang2013robust, fan2017concordance}. But when learning decision rules online, we are often faced with the classic dilemma between exploration and exploitation. That is, we will either exploit our current knowledge and possibly fail to learn something new or explore actions with uncertain rewards and possibly receive fewer rewards than the currently best decision.
    
Since the formulation of the contextual bandit problem, a considerable amount of variants have been studied and the corresponding solutions are offered. Readers are referred to \citet{tewari2017ads} for a comprehensive survey of solutions including parametric and non-parametric learning methods. The techniques to balance the exploration-and-exploitation trade-off are categorized by \citet{kaelbling1996reinforcement} as greedy strategies, randomized strategies, and interval-based strategies. The first kind of strategies always choose the arm with the highest expected reward, and \citet{bastani2017exploiting} showed that consistent estimation of the reward model parameters can be achieved under strong assumptions. Randomized strategies will choose the best arm by default and explore random actions with a small probability. Interval-based strategies are extensively studied by computer science researchers and there are two main branches: the Upper Confidence Bound methods which choose the action with the highest upper bound of reward \citep[see][]{auer2002using, dani2008stochastic, chu2011contextual}, and the Action Elimination methods which eliminate actions whose upper bound is lower than the expected reward of the best action \citep[see][]{perchet2013multi, qian2016randomized}. Besides those three main categories, \citet{agrawal2013thompson} used Thompson sampling to induce exploration, \citet{caelen2007improving} introduced the notion of probability of correct selection to improve the exploration step. We will focus on randomized strategies since they can be easily implemented in practice and their statistical properties including inference can be rigorously studied.
    
There are two main varieties of randomized strategies in antecedent literature and one of them is the ``certainty equivalence with forcing'' policy, which takes random action at predetermined decision time and the current optimal action at other time. \citet{goldenshluger2013linear} adopted this policy to solve the contextual bandit problem under the linear model setting. Their strategy maintains two sets of samples, of which one is independent, so that the consistency of the parameter estimators can be established. \citet{bastani2015online} further studied the linear contextual bandit problem in the high-dimensional case, and the same kind of strategy was used. A more commonly used and simpler strategy to handle the exploration-and-exploitation problem is the $\varepsilon$-greedy policy, which chooses the current optimal action with probability $1-\varepsilon$ and makes a random decision with probability $\varepsilon$. \citet{chambaz2017targeted} studied the asymptotic properties of the reward under a general parametric model using such a policy with $\varepsilon$ being a function of the estimated treatment effects. \citet{yang2002randomized} and \citet{qian2016kernel} also used the $\varepsilon$-greedy policy, but their $\varepsilon$ is a function of decision time. They used non-parametric methods to estimate the reward and proved that the estimated decision rule is consistent if $\varepsilon$ decreases at certain rates. Our work will fill in the blank of the combination of the linear reward model setting and $\varepsilon$-greedy policy.

Our contributions are from several folds. First, all the contextual bandit literature mentioned above focuses on the analysis of the regret, the difference between the cumulative rewards following the learning method and that following the oracle policy. But none of them studied the asymptotic distribution of the parameter estimators, which is an important inference problem for online decision as \citet{tewari2017ads} discussed in their survey that work should be done to assess the ``usefulness of contextual variables''. When the asymptotic distributions are known, we can easily test the significance of features, and it offers a traditional way to select variables. In the present paper, we solve the two-armed contextual bandit problem under the linear model setting using the $\varepsilon$-greedy policy. We prove the consistency of the online ordinary least squares (OLS) estimator under certain conditions of the exploration parameter $\varepsilon$ and find its asymptotic distribution by the martingale central limit theorem. We also propose an in-sample inverse propensity weighted (IPW) estimator for the value function of the derived online recommendation rule. Based on the asymptotic properties of the parameter estimators, we are able to establish the asymptotic normality of the IPW value estimator.
    
Second, another topic missing from the linear bandit literature is model misspecification \citep{dimakopoulou2017estimation}. When the linear model is misspecified, we expect the estimator to converge to the least false parameter, as is proved by \citet{white1980using} for the regression case without online decision-making. However, it is not true for the online decision-making OLS estimator under the $\varepsilon$-greedy policy. Because any greedy or soft-greedy policy will cause sampling bias and hence biased estimator of the least false parameter. Therefore we propose an online weighted least squares (WLS) estimator to correct the sampling bias and show its asymptotic properties. The IPW value estimator based on the online WLS estimator is also shown to be asymptotically normal.
    

The rest of the paper is organized as follows. We introduce the model notation and present a general problem setting and a learning method for the correctly specified model in Section \ref{sec:problem_setting_and_decision_policy}. Main results for the online OLS estimator and the IPW value estimator under the correct linear model are given in Section \ref{sec:asymptotic_properties_of_the_online_OLS_estimator}. We discuss the problem arising from model misspecification and propose the WLS estimator as a remedy in Section \ref{sec:misspecified_model}. Numerical results on simulated data and Yahoo! news article recommendation data are presented in Sections \ref{sec:numerical_study} and \ref{sec:yahoo} respectively. We briefly discuss the regret bound of the algorithm and some possible extensions of our work in Section \ref{sec:discussions}. Proofs of main results and auxiliary results are presented in the online supplement.
    
\section{Problem Setting and Decision Policy}\label{sec:problem_setting_and_decision_policy}
Suppose we need to make decisions at a sequence of time points, $\mathcal{T}=\{1,\cdots, T\}$, and at each time point $t\in \mathcal{T}$ we observe the i.i.d. contextual information $\V{z}_t\in\R^{d-1}$. Denote $\V{x}_t = (1, \V{z}_t^T)^T$ and its joint distribution as $\mathcal{P}_\V{x}$. After taking an action $a_t$ from the possible action set $\mathcal{A}=\{0,1\}$, we can observe the reward $y_t$. We assume larger reward is preferable, and the underlying true model for $y_t$ is
\begin{equation}\label{eq:model}
    y_t=a_t\V{x}_t^T\V{\beta}_1+(1-a_t)\V{x}_t^T\V{\beta}_0+e_t,
\end{equation}
where $\V{\beta}_0, \V{\beta}_1 \in \mathbb{R}^d$ are the model parameters, $e_t$ conditional on $a_t$ are i.i.d. from a $\sigma$-subgaussian distribution \footnote{A random variable $X$ is $\sigma$-subgaussian if $\E(\exp\{cX\})\le \exp\{c^2\sigma^2/2\}$ for all $c\in\R$. It implies that $\E(X) = 0$ and $\E(X^2) < \sigma^2$.} and $e_t\perp \V{x}_t|a_t$ for all $t\in \mathcal{T}$. Denote the conditional distribution of $e_t|a_t=i$ to be $\mathcal{P}_{ei}$ for $i=0,1$. 
    
There are two kinds of online decision policies in terms of their relationships to previous data: random policies that are totally independent of historical information; and dependent policies that make use of previous information in the hope of maximizing the reward. One simple dependent policy is the greedy policy, which will always choose the action yielding the highest estimated reward. In some cases, however, the greedy policy may be trapped in some actions that have good performance in the early stage and leave other actions unexplored. Therefore, we need to deal with the exploitation and exploration trade-off, and a commonly adopted adaption to the greedy policy is the $\varepsilon$-greedy policy. To apply it, we need to first learn a model from the warm-up samples following the random policy and update the model afterward. Suppose we have a small amount of data following the random policy before $T_0+1$. We posit the same model as the true one in (\ref{eq:model}) and fit it using OLS to get $\hat{\V{\beta}}_{0,T_0}$ and $\hat{\V{\beta}}_{1,T_0}$. Then we are able to define the decision rule under the $\varepsilon$-greedy policy and update the model recurrently. For $t=T_0+1,T_0+2,\cdots, T$, define
\begin{equation}\label{eq:pi}
    \hat{\pi}_t=(1-\varepsilon_t)I\left\{(\hat{\V{\beta}}_{1,t-1}-\hat{\V{\beta}}_{0,t-1})^T\V{x}_t\ge0\right\}+\frac{\varepsilon_t}{2},
\end{equation}
then the stochastic decision is
\begin{equation}\label{eq:a}
    \hat{a}_t\sim \mathrm{Bernoulli}(\hat{\pi}_t).
\end{equation}
Here $\varepsilon_t$ is a function of $t$ bounded between 0 and 1. The smaller $\varepsilon_t$ is, the more choices are made according to our greedy policy, and hence the higher return in the long run. But the learning rate is slower at the early stage if $\varepsilon_t$ is small. So usually a decreasing function is favorable. Let $\hat{\pi}_t=1/2$ when $t\le T_0$, and $\hat{a}_t\sim \mathrm{Bernoulli}(\hat{\pi}_t)$ become the decisions of the balanced random policy. Then for $t=T_0+1,T_0+2,\cdots, T$, the online OLS estimator is
\begin{equation}\label{eq:betaest}
    \hat{\V{\beta}}_{i,t}=\left(\sum_{s=1}^tI\{\hat{a}_s=i\}\V{x}_s\V{x}_s^T\right)^{-1}\sum_{s=1}^tI\{\hat{a}_s=i\}\V{x}_sy_s, \; i=0,1,
\end{equation}
if the sample second moment $\sum_{s=1}^t I\{\hat{a}_s=i\}\V{x}_s\V{x}_s^T$ is invertible for each step $t$. The above definitions of the decision rule and the online OLS estimator depend on each other: the $\varepsilon$-greedy decision at time $t$ is made with previous estimation of the model, then the following reward $y_{t}$ will depend on $\hat{a}_t$ through the true model (\ref{eq:model}), and exert influence on the next estimation. Since the data are not independent, the consistency and asymptotic normality of the estimator do not follow from the Law of Large Numbers and Central Limit Theorem easily.
    
\section{Inference Under Correctly Specified Model}\label{sec:asymptotic_properties_of_the_online_OLS_estimator}
    
\subsection{Parameter Inference}
We will first give the consistency of the online OLS estimator. \citet{bastani2015online} showed that the tail bound of the online OLS estimator is controlled by the proportion of the forced independent samples. So it suffices to show that our $\varepsilon$-greedy policy draws independent samples frequently enough such that the consistency is guaranteed. 
    
Denote the sigma field generated by the history information up to time $t$ as 
$$
\mathcal{F}_t=\sigma\langle \V{x}_1, \hat{a}_1, y_1,\cdots, \V{x}_t, \hat{a}_t, y_t \rangle.
$$ 
The estimated parameters for both actions have the same structure. Consider the estimator for the parameters associated with action 1, where $I\{\hat{a}_s=1\}=\hat{a}_s$, we have
$$
\hat{\V{\beta}}_{1,t}-\V{\beta}_1=\left(\frac{1}{t}\sum_{s=1}^t\hat{a}_s\V{x}_s\V{x}_s^T\right)^{-1}\frac{1}{t}\sum_{s=1}^t\hat{a}_s\V{x}_se_s
$$
from (\ref{eq:betaest}). Note that the second summation $\sum_{s=1}^t\hat{a}_s\V{x}_se_s$ is a martingale wrt $\{\mathcal{F}_s\}_{s=1}^t$ since 
$$
\E(\hat{a}_s\V{x}_se_s|\mathcal{F}_{s-1})=\E[\E(\hat{a}_s\V{x}_se_s|\mathcal{F}_{s-1},\V{x}_s)|\mathcal{F}_{s-1}]=0.
$$
We want to bound the mean of the martingale differences using concentration inequalities and prevent the first inverse term from blowing up. The following two assumptions on the distribution of $\V{x}_t$ are needed. 
    
\begin{assumption}\label{as:1}
    $\mathcal{P}_\V{x}$ is dominated by the Lebesgue measure on $\mathbb{R}^d$. There exists a positive constant $L_x$ such that $\lVert \V{x} \rVert_\infty \le L_x$ for all realizations of $\V{x}\sim\mathcal{P}_\V{x}$.
\end{assumption}
    
\begin{assumption}\label{as:2}
    $\Sigma=\E_{\V{x}\sim\mathcal{P}_\V{x}}(\V{x}\V{x}^T)$ has minimum eigenvalue $\lambda_{\min}(\Sigma)>\lambda$ for some $\lambda>0$.
\end{assumption}
    
Assumption \ref{as:1} ensures that the mean of the martingale differences will converge to zero. Assumption \ref{as:2} guarantees that the sample second moment $\hat{\Sigma}=\frac{1}{t}\sum_{s=1}^t\V{x}_s\V{x}_s^T$ is non-singular and hence $\hat{\V{\beta}}_{i,t}$ exists with high probability. The only term we have not controlled is the decision $\hat{a}_s$. In the worst scenario where one action is always superior to the other, all decisions under the greedy policy will take the better one and only the random decisions may choose the other one with probability $\varepsilon_t/2$. If $\varepsilon_t$ is a decreasing function, we cannot ensure that the first summation $\sum_{s=1}^t\hat{a}_s\V{x}_s\V{x}_s^T$ still grows linearly in $t$. Fortunately, we find out that the minimum eigenvalue of the first summation only needs to grow faster than $\sqrt{t}$. Then a careful choice of $\varepsilon_t$ can meet the requirement for the consistency of the estimator.
    
\begin{restatable}[Tail bound for the online OLS estimator]{proposition}{olstail}\label{prop:tail}
    In the online decision-making model with the $\varepsilon$-greedy policy, if Assumptions \ref{as:1} and \ref{as:2} are satisfied, and $\varepsilon_t$ is non-increasing, then for any $\varkappa>0$,
    \begin{equation*}
        \begin{split}
            P\left(\|\hat{\V{\beta}}_{i,t}-\V{\beta}_i\|_1\le\varkappa\right) \ge &1-\exp\left\{-\frac{t\varepsilon_t}{8}\right\}-d\exp\left\{-\frac{t\varepsilon_t\lambda}{32L_x^2}\right\}-2d\exp\left\{-\frac{t\varepsilon_t^2\lambda^2\varkappa^2}{128d^2\sigma^2L_x^2}\right\}\\ &\;\, +2d^2\exp\left\{-\frac{t\varepsilon_t^2\lambda^2\varkappa^2 + 4t\varepsilon_t\lambda d^2\sigma^2}{128d^2\sigma^2L_x^2}\right\}, \; i=0,1.
        \end{split}
    \end{equation*}
\end{restatable}
    
Proposition \ref{prop:tail} states that if $t\varepsilon_t^2\to\infty$ as $t\to\infty$, then the probability of $\|\hat{\V{\beta}}_{i,t}-\V{\beta}_i\|_1\le\varkappa$ goes to $1$ for any $\varkappa>0$. Any constant $\varepsilon_t$ satisfies this trivially.  Moreover, some decreasing examples include $\varepsilon_t = kt^{-0.499}$, $k\log t/\sqrt{t}$, or $k\log\log t/\sqrt{t}$ with some constant $k$. For a finite $t$, the probability of $\|\hat{\V{\beta}}_{i,t}-\V{\beta}_i\|_1\le\varkappa$ is smaller for bigger covariates dimension $d$, reward standard deviation $\sigma$, covariates bound $L_x$, and/or smaller $\lambda$. This means the convergence of the estimator may take longer time if we have high dimensional features, or the sample rewards are noisy. Covariates should be rescaled before fitting the model such that $L_x$ is small and collinearity should be checked to prevent singularity.

The consistency of $\hat{\V{\beta}}_{i,t}$ then follows from Proposition \ref{prop:tail} trivially. 
    
\begin{restatable}[Consistency of the online OLS estimator]{corollary}{olsconsis}\label{col:1}
    If Assumptions \ref{as:1} and \ref{as:2} are satisfied, $\varepsilon_t$ is non-increasing and $t\varepsilon_t^2\to\infty$ as $t\to\infty$, then the online OLS estimator $\hat{\V{\beta}}_{i,t}$ is a consistent estimator for $\V{\beta}_i$, $i=0,1$.
\end{restatable}
    
The asymptotic normality of the online parameter estimator can be established using the Martingale Central Limit Theorem \citep[see Theorem 3.2 in][]{hall1980martingale}.
    
\begin{restatable}[Asymptotic normality of the online OLS estimator]{theorem}{olsnorm}\label{thm:1}
    In the online decision-making model with $\varepsilon$-greedy policy, if Assumptions \ref{as:1} and \ref{as:2} are satisfied, the non-increasing function $\varepsilon_t$ has $\varepsilon_t\to\varepsilon_\infty$ and $t\varepsilon_t^2\to\infty$ as $t\to\infty$. Then
    \begin{equation*}
        \begin{split}
            \sqrt{t}\left(\hat{\V{\beta}}_{i,t}-\V{\beta}_i\right)\overset{d}{\to} \mathcal{N}_d\left(\V{0},S_i\right), \; i=0,1,
        \end{split}
    \end{equation*}
    where 
    \begin{equation}\label{eq:var}
        S_i = \sigma_i^2\left(\frac{\varepsilon_\infty}{2}\int_{\mathcal{X}_0\cup\mathcal{X}_1} \V{x}\V{x}^T d\mathcal{P}_\V{x}+(1-\varepsilon_\infty)\int_{\mathcal{X}_i} \V{x}\V{x}^T d\mathcal{P}_\V{x}\right)^{-1} 
    \end{equation}
    with $\mathcal{X}_0 = \{\V{x}:(\V{\beta}_1-\V{\beta}_0)^T\V{x}< 0\}$, $\mathcal{X}_1 = \{\V{x}:(\V{\beta}_1-\V{\beta}_0)^T\V{x}\ge 0\}$, and $\sigma_i^2=\E(e_s^2|a_s=i)$. A consistent estimator for $S_i$ is given by
    $$
    \frac{\sum_{s=1}^t I\{\hat{a}_s=i\}\hat{e}_s^2}{\sum_{s=1}^t I\{\hat{a}_s=i\}}\left(\frac{1}{t}\sum_{s=1}^{t}I\{\hat{a}_s=i\}\V{x}_s\V{x}_s^T\right)^{-1},
    $$
    where $\hat{e}_s=y_s-\hat{\V{\beta}}_{i,t}^T\V{x}_s$.
\end{restatable}

As shown in (\ref{eq:var}), the asymptotic variance of the online OLS estimator depends on the limit of $\varepsilon_t$. If $\varepsilon_t = \varepsilon_\infty = 1$, then the result is the same as that from a randomized experiment. Note that 
$$
\sigma^2_0 S_0^{-1} + \sigma^2_1 S_1^{-1} = \int_{\mathcal{X}_0\cup\mathcal{X}_1} \V{x}\V{x}^T d\mathcal{P}_\V{x}
$$
is a fixed matrix for a given $\mathcal{P}_{\V{x}}$. So we cannot choose a $\varepsilon_\infty$ to minimize the asymptotic variance of both $\hat{\V{\beta}}_{0,t}$ and $\hat{\V{\beta}}_{1,t}$ at the same time, yet it is possible to adjust $\varepsilon_\infty$ to get a more efficient estimator for one action only if we have prior knowledge of $\mathcal{P}_\V{x}$. In practice, we suggest setting $\varepsilon_\infty = 0$ so that we could fully exploit the model and achieve higher rewards in the long term. In that case, the asymptotic variance of $\hat{\V{\beta}}_{i,t}$ is not affected by the variance of the design distribution in the other subspace.

The difference of the two 
estimators $\hat{\V{\beta}}_{1,t}-\hat{\V{\beta}}_{0,t}$ can also be written as the summation of martingale differences. Following similar strategies, it is easy to derive its asymptotic distribution, i.e.
$$\sqrt{t}\left(\left(\hat{\bm{\beta}}_{1,t} - \hat{\bm{\beta}}_{0,t}\right) - (\bm{\beta}_1 - \bm{\beta}_0)\right)\overset{d}{\to} \mathcal{N}_d\left(\V{0},S_0 + S_1\right).$$
This implies that the asymptotic covariance between $\hat{\V{\beta}}_{0,t}$ and $\hat{\V{\beta}}_{1,t}$ is zero. An intuition for this result is because there is no overlap between the data used for estimating $\bm{\beta}_0$ and $\bm{\beta}_1$. Thus, the variance of $\hat{\V{\beta}}_{1,t}-\hat{\V{\beta}}_{0,t}$ can be estimated by the sum of the variance estimates of $\hat{\V{\beta}}_{0,t}$ and $\hat{\V{\beta}}_{1,t}$, respectively.
    
\subsection{Value Inference}
Given the true underlying reward model, the best possible policy is the greedy policy based on the true expected reward. In the linear reward model case, it will assign action $a_t =  I\{\V{\beta}_1 - \V{\beta}_0)^T \V{x}_t \ge 0 \}$ at each decision time $t$. We call such a policy the oracle policy and define the expected value under the oracle policy as the optimal value
\begin{equation}\label{eq:opt_value}
    V = \int_{(\V{\beta}_1 - \V{\beta}_0)^T \V{x} \ge 0 }\V{\beta}_1^T \V{x}d\mathcal{P}_\V{x} + \int_{(\V{\beta}_1 - \V{\beta}_0)^T \V{x} < 0 }\V{\beta}_0^T \V{x}d\mathcal{P}_\V{x}.
\end{equation}
In our proposed online decision-making process, the expected value under the estimated $\varepsilon$-greedy policy at time $t$ (indexed by $\hat{\pi}$) is 
\begin{equation}\label{eq:est_value}
    V_t^{\hat{\pi}} = \int \hat{a}\V{\beta}_1^T \V{x}d\mathcal{P} + \int (1 - \hat{a})\V{\beta}_0^T \V{x}d\mathcal{P}
\end{equation}
with
\begin{equation}\label{eq:va}
    \hat{a}\sim \mathrm{Bernoulli}\left(\hat{\pi}=(1-\varepsilon_t)I\left\{(\hat{\V{\beta}}_{1,t-1}-\hat{\V{\beta}}_{0,t-1})^T\V{x}\ge0\right\}+\frac{\varepsilon_t}{2}\right),
\end{equation}
where $\mathcal{P}$ is the joint distribution of $\V{x}$ and $\hat{a}$. By definition, the expected value $V_t^{\hat{\pi}}$ is an out-of-sample evaluation of the performance of the current estimated policy. Note that $V^{\hat{\pi}}_t$ is a continuous function of $\hat{\V{\beta}}_{0, t - 1}$, $\hat{\V{\beta}}_{1, t - 1}$ and $\varepsilon_t$. By continuous mapping theorem, it converges in probability to the optimal value $V$ since $\hat{\V{\beta}}_{0, t}$ and $\hat{\V{\beta}}_{1, t}$ converge to $\V{\beta}_0$ and $\V{\beta}_1$ by Corollary 3.1. Thus the optimal value gives us a sense of how far we can go under the $\varepsilon$-greedy policy and it can be used to assess how well the current estimated policy performs. But the true model and the distribution of $\V{x}$ is unknown in practice, so the optimal value $V$ has to be estimated. To that end, we propose an IPW estimator
\begin{equation}\label{eq:ipw}
    \hat{V}_t = \frac{1}{t}\sum_{s=1}^t \frac{I\{\hat{a}_s = I\{(\hat{\V{\beta}}_{1,s-1} - \hat{\V{\beta}}_{0,s-1})^T\V{x}_s \ge 0\}\}}{P\{\hat{a}_s = I\{(\hat{\V{\beta}}_{1,s-1} - \hat{\V{\beta}}_{0,s-1})^T\V{x}_s \ge 0\}\}}y_s,
\end{equation}
where we set $\hat{\V{\beta}}_{1,s-1} - \hat{\V{\beta}}_{0,s-1} = 0$ for $1 \le s \le T_0$ so that the definition is consistent with our decision rule. In contrast to the expected value under the estimated policy $V_t^{\hat{\pi}}$, this IPW estimator is an in-sample estimator that does not require the knowledge of the true model or the covariates distribution. It corrects the bias from the random exploration by the inverse propensity weighting and is hence an unbiased estimator of the optimal value $V$. Furthermore, its asymptotic normality can be shown under the following extra assumption.
    
\begin{assumption}\label{as:4}
    There exists $C > 0$, such that for $\V{x}\sim\mathcal{P}_\V{x}$,
    $$
    P(0<|(\V{\beta}_1 - \V{\beta}_0)^T \V{x}| \le \ell) \le C \ell, \; \forall \ell > 0.
    $$
\end{assumption}
    
Assumption \ref{as:4} is a margin condition widely adopted by contextual bandit literature \citep{goldenshluger2013linear, bastani2015online}. Intuitively, when feature covariates are near the decision boundary $(\V{\beta}_1 - \V{\beta}_0)^T \V{x} = 0$, estimated policies are more likely to make wrong decisions because the rewards are similar for both actions. Assumption \ref{as:4} simplifies our problem by restricting the probability of having such covariates and letting the variance caused by the wrong decisions go away. In the end, we can achieve a clear and estimable variance of the value estimator with its help. We note that, however, it is not a necessary assumption for the asymptotic normality of the value estimator. More discussions are deferred to Section \ref{sec:misspecified_model} where a more general case is considered.
    
\begin{restatable}[Asymptotic normality of the IPW value estimator under correctly specified models]{theorem}{vnorm}\label{thm:3}
    In the online decision-making model with the $\varepsilon$-greedy policy, if Assumptions \ref{as:1}--\ref{as:4} are satisfied, the non-increasing function $\varepsilon_t$ satisfies $\varepsilon_t\to\varepsilon_\infty$ and $t\varepsilon_t^2\to\infty$ as $t\to\infty$. Then
    \begin{equation*}
        \sqrt{t}\left(\hat{V}_t-V\right)\overset{d}{\to} \mathcal{N}(0,\zeta^2),
    \end{equation*}
    where 
    \begin{align*}
        \zeta^2 =& \frac{2}{2-\varepsilon_\infty}\left[\int_{\V{\beta}^T \V{x} \ge 0}\{(\V{\beta}_1^T \V{x})^2+\sigma_1^2\} d\mathcal{P}_\V{x} + \int_{\V{\beta}^T \V{x} < 0}\{(\V{\beta}_0^T \V{x})^2+\sigma_0^2\} d\mathcal{P}_\V{x}\right]\\
        &-\left(\int_{\V{\beta}^T \V{x} \ge 0}\V{\beta}_1^T \V{x}d\mathcal{P}_\V{x} + \int_{\V{\beta}^T \V{x} < 0}\V{\beta}_0^T \V{x}d\mathcal{P}_\V{x}\right)^2
    \end{align*}
    and $\sigma_i^2=\E(e_s^2|a_s=i)$. A consistent estimator for $\zeta^2$ is given by
    \begin{equation*}
        \hat{\zeta}_t^2 =\frac{2}{2-\varepsilon_t}\frac{1}{t}\sum_{s=1}^t \frac{I\{\hat{a}_s = I\{\hat{\V{\beta}}_{s-1}^T\V{x}_s \ge 0\}\}}{P\{\hat{a}_s = I\{\hat{\V{\beta}}_{s-1}^T\V{x}_s \ge 0\}\}}y_s^2 - \hat{V}_t^2.
    \end{equation*}
    
\end{restatable}

\section{Extension to Model Misspecification}\label{sec:misspecified_model}
    
\subsection{Model Setting}
In the online decision-making model, assume the true underlying model of $y_t$ is
\begin{equation}\label{eq:mis-model}
    y_t=a_t\phi_1(\V{x}_t)+(1-a_t)\phi_0(\V{x}_t)+e_t,
\end{equation}
where $\phi_0$ and $\phi_1$ are unknown arbitrary measurable functions of $\V{x}_t$, and $\V{x}_t$ and $e_t$ satisfy the same conditions as before. In order to fit the data, we may still posit the linear model in (\ref{eq:model}), and then the true model can be written as
\begin{equation}
    y_t=a_t\V{x}_t^T\V{\beta}_1+(1-a_t)\V{x}_t^T\V{\beta}_0+u_t,
\end{equation}
where $u_t\equiv a_t\{\phi_1(\V{x}_t)-\V{x}_t^T\V{\beta}_1\}+(1-a_t)\{\phi_0(\V{x}_t)-\V{x}_t^T\V{\beta}_0\}+e_t$. \citet{white1980using} showed that the OLS estimator under such model misspecification will converge to the least false parameter, the minimizer of the error from both the random noise and misspecification, that is,
\begin{equation}\label{eq:leastfalse}
    \V{\beta}_i^*=\arg\min_{\V{\beta}_i} \int \{\phi_i(\V{x})-\V{x}^T\V{\beta}_i\}^2 d\mathcal{P}_\V{x}+\sigma_i^2, \; i=0,1.
\end{equation}
It can be found by setting the derivative of the objective function to 0 that
$$
\V{\beta}_i^*=\left(\int \V{x}\V{x}^T d\mathcal{P}_\V{x}\right)^{-1}\int \phi_i(\V{x})\V{x} d\mathcal{P}_\V{x}, \; i=0,1.
$$
Unlike the OLS estimator under the i.i.d. offline setting, the online OLS estimator will no longer converge to the least false parameter if we follow the $\varepsilon$-greedy decision policy. Intuitively, if the true model is not linear, the slopes will be different at different values of $\V{x}_t$. So the data sampled under the random policy and greedy policy will lead to different slope estimations. Here the definition of the least false parameter is based on covariates from the population, which is comparable to data sampled under the random policy by Law of Large Numbers, while the online OLS estimator is based on data sampled under the $\varepsilon$-greedy policy. Therefore the online OLS estimator under our policy will be biased for the least false parameter. Fortunately, this bias from sampling can be corrected by weighting the samples. The proposed online WLS estimator is defined as follows. Let $\tilde{\V{\beta}}_i(T_0)=\hat{\V{\beta}}_i(T_0),\; i=0,1$ and $\tilde{\pi}_t=1/2$ for $t\le T_0$. For $t=T_0+1, T_0+2,\cdots, T$, define $\tilde{\pi}_t$ the same as (\ref{eq:pi}) but substitute $\hat{\V{\beta}}_{i,t}$ with $\tilde{\V{\beta}}_{i,t}$. For all $t$, $\tilde{a}_t$ is the same as (\ref{eq:a}) with $\tilde{\pi}_t$ in place of $\hat{\pi}_t$. For $t=T_0+1, T_0+2,\cdots, T$,
\begin{equation}
    \tilde{\V{\beta}}_{i,t}=\left(\sum_{s=1}^t\frac{I\{\tilde{a}_s=i\}}{P(\tilde{a}_s=i)}\V{x}_s\V{x}_s^T\right)^{-1}\sum_{s=1}^t\frac{I\{\tilde{a}_s=i\}}{P(\tilde{a}_s=i)}\V{x}_s y_s,\; i=0,1.
\end{equation}
Note that $P(\tilde{a}_s=1)=\tilde{\pi}_s$ and $P(\tilde{a}_s=0)=1-\tilde{\pi}_s$, which could be or be very close to $0$ if $\varepsilon_s$ is or converges to $0$. Then the estimator could be very unstable if some samples have infinite weights. To avoid this, we will need $\varepsilon_t>0$ for all $t$. The simplest case is setting $\varepsilon_t$ to a constant, but we could also set $\varepsilon_t$ as a non-increasing function of $t$ with its limit greater than 0 when $t$ goes to infinity.
    
\subsection{Parameter Inference}
Again, we rely on the Martingale structure of $\tilde{\V{\beta}}_{i,t} - \V{\beta}_i^*$ to prove the consistency and asymptotic normality of the online WLS estimator. We need the following assumption to control the new error term $u_t$.
    
\begin{assumption}\label{as:5} Linear approximation of the true model is reliable, that is, there exists $L_\phi>0$ such that
    $$
    |\phi_i(\V{x})-\V{x}^T\V{\beta}^*_i|\le L_\phi, \; i = 0,1,
    $$
    for all realizations of $\V{x}\sim\mathcal{P}_\V{x}$.
\end{assumption}
    
Given Assumption \ref{as:1}, what Assumption \ref{as:5} really constrains is that $\phi_i(\V{x})$ should be bounded for $\V{x} \in [-L_x, L_x]^d$. For example, $\exp(\V{x}^T\V{\beta})$,  $1/(1+\exp(-\V{x}^T\V{\beta}))$, and many other common functions satisfy this assumption. Under Assumption \ref{as:5}, the mean of martingale differences will converge to 0 by an concentration inequality (Lemma 1 in the online supplement). The next proposition establishes the tail bound for the online WLS estimator.
    
\begin{restatable}[Tail Bound for the online WLS estimator]{proposition}{wlsconsis}\label{prop:tail-mis}
    In the online decision-making model with the $\varepsilon$-greedy policy, if Assumptions \ref{as:1}, \ref{as:2} and \ref{as:5} are satisfied, and $\varepsilon_t$ is non-increasing with $\varepsilon_t\to\varepsilon_\infty>0$ as $t\to\infty$. Then for any $\varkappa>0$,
    \begin{equation*}
        \begin{split}
            P\left(\|\tilde{\V{\beta}}_{i,t}-\V{\beta}_i^*\|_1\le\varkappa\right) \ge 1&-\exp\left\{-\frac{t\varepsilon_t}{8}\right\}-d\exp\left\{-\frac{t\varepsilon_t\lambda}{32L_x^2}\right\}-2d\exp\left\{-\frac{t\varepsilon_t^4\lambda^2\varkappa^2}{512d^2L_x^2L_\phi^2}\right\}\\&-2d\exp\left\{-\frac{t\varepsilon_t^2\lambda^2\varkappa^2}{512d^2\sigma^2L_x^2}\right\}+2d^2\exp\left\{-\frac{t\varepsilon_t^4\lambda^2\varkappa^2+8t\varepsilon_t\lambda d^2L_\phi^2}{512d^2L_x^2L_\phi^2}\right\}\\&+2d^2\exp\left\{-\frac{t\varepsilon_t^2\lambda^2\varkappa^2+8t\varepsilon_t\lambda d^2\sigma^2}{512d^2\sigma^2L_x^2}\right\},\; i=0,1. 
        \end{split}
    \end{equation*}
\end{restatable}
    
The tail bound has two more terms than that given in Proposition \ref{prop:tail} due to model misspecification. A bigger misspecification bound $L_\phi$ would lead to slower convergence. Other constants including $d$, $\sigma$, $L_x$ and $\lambda$ play the same role as in Proposition \ref{prop:tail}. When $t$ goes to infinity, all exponential terms go to zero since $\varepsilon_\infty>0$, the consistency then follows without extra conditions.
    
\begin{corollary}[Consistency of the online WLS estimator]\label{col:2}
    If Assumptions \ref{as:1}, \ref{as:2} and \ref{as:5} are satisfied, $\varepsilon_t$ is non-increasing and $\varepsilon_t\to\varepsilon_\infty>0$ as $t\to\infty$, then the online WLS estimator $\tilde{\V{\beta}}_{i,t}$ is a consistent estimator for the least false parameter $\V{\beta}_i^*$, $i=0,1$.
\end{corollary}
    
Then the asymptotic normality can be established using the Martingale Central Limit Theorem.
    
\begin{restatable}[Asymptotic Normality of the online WLS estimator]{theorem}{wlsnorm}\label{thm:2}
    In the online decision-making model with the $\varepsilon$-greedy policy, if Assumptions \ref{as:1}, \ref{as:2} and \ref{as:5} are satisfied, $\varepsilon_t$ is non-increasing and $\varepsilon_t\to\varepsilon_\infty>0$ as $t\to\infty$. Then
    $$
    \sqrt{t}\left(\tilde{\V{\beta}}_{i,t}-\V{\beta}_i^*\right)\overset{d}{\to} \mathcal{N}_d\left(\V{0},\Sigma^{-1}H_i\Sigma^{-1}\right), \; i=0,1,
    $$
    where 
    $$
    H_i=\int \frac{\V{x}\left(\phi_i(\V{x})-\V{x}^T\V{\beta}^*_i\right)^2\V{x}^T+\sigma_i^2\V{x}\V{x}^T}{(1-\varepsilon_\infty)\left(iI\{\V{\beta}_1^{*T}\V{x}\ge \V{\beta}_0^{*T}\V{x}\} + (1-i)I\{\V{\beta}_1^{*T}\V{x}< \V{\beta}_0^{*T}\V{x}\}\right)+\varepsilon_\infty/2}d\mathcal{P}_\V{x}.
    $$
    and $\sigma_i^2=\E(e_s^2|a_s=i)$. A consistent estimator for $\Sigma^{-1}H_i\Sigma^{-1}$ is given by $\hat{\Sigma}_{t}^{-1}\hat{H}_{i, t}\hat{\Sigma}_{t}^{-1}$ with 
    $$
    \hat{\Sigma}_{t} = \frac{1}{t}\sum_{s=1}^{t}\V{x}_s\V{x}_s^T
    $$
    and
    $$
    \hat{H}_{i, t} = \frac{1}{t}\sum_{s=1}^t \frac{I\{\tilde{a}_s=i\}}{\{P(\tilde{a}_s=i)\}^2}\V{x}_s(y_s-\V{x}_s^T\tilde{\V{\beta}}_{i,t})^2\V{x}_s^T.
    $$
\end{restatable}
    
The consistent estimator of the asymptotic variance takes a sandwich form. It has a similar structure as the estimated variance of the WLS estimator under an independent offline setting.
    
\subsection{Value Inference}
If the model is misspecified with true reward model (\ref{eq:mis-model}), define the optimal value as
\begin{equation}\label{eq:opt_value_mis}
    V^* = \int_{(\V{\beta}_1^* - \V{\beta}_0^*)^T \V{x} \ge 0 }\phi_1(\V{x})d\mathcal{P}_\V{x} + \int_{(\V{\beta}_1^* - \V{\beta}_0^*)^T \V{x} < 0 }\phi_0(\V{x})d\mathcal{P}_\V{x},
\end{equation}
which is the expected value under the best possible policy based on the linear model. Then the expected value under the estimated policy at time $t$ is 
\begin{equation}\label{eq:est_value_mis}
    V_t^{\tilde{\pi}} = \int \tilde{a}\phi_1(\V{x})d\mathcal{P} + \int (1 - \tilde{a})\phi_0(\V{x})d\mathcal{P}
\end{equation}
with
\begin{equation}\label{eq:va_mis}
    \tilde{a}\sim \mathrm{Bernoulli}\left(\tilde{\pi}=(1-\varepsilon_t)I\left\{(\tilde{\V{\beta}}_{1,t-1}-\tilde{\V{\beta}}_{0,t-1})^T\V{x}\ge0\right\}+\frac{\varepsilon_t}{2}\right),
\end{equation}
where $\mathcal{P}$ is the joint distribution of $\V{x}$ and $\tilde{a}$. Since the consistency of the WLS estimator requires the limit of $\varepsilon_t$ to be bounded away from zero, there is always a gap between $V_t^{\tilde{\pi}}$ and $V^*$. To estimate the optimal value, we still consider the IPW estimator
$$
\tilde{V}_t = \frac{1}{t}\sum_{s=1}^t \frac{I\{\tilde{a}_s = I\{(\tilde{\V{\beta}}_{1,s-1} - \tilde{\V{\beta}}_{0,s-1})^T\V{x}_s \ge 0\}\}}{P\{\tilde{a}_s = I\{(\tilde{\V{\beta}}_{1,s-1} - \tilde{\V{\beta}}_{0,s-1})^T\V{x}_s \ge 0\}\}}y_s,
$$
where we set $\tilde{\V{\beta}}_{1,s-1} - \tilde{\V{\beta}}_{0,s-1} = 0$ for $1\le s \le T_0$ to be consistent with our decision rule. It is still unbiased for the optimal value but its asymptotic variance is more complicated than the situation we discussed in Section \ref{sec:asymptotic_properties_of_the_online_OLS_estimator}. Previously when the model is correctly specified, wrong decisions may occur when the rewards from both actions are similar. Now, more decisions can be wrong because they are based on the linear projection of the true reward model, which will further complicate the estimation variation. So the variance introduced by wrong decisions cannot be handled by a simple assumption like \ref{as:4}. In order to establish the asymptotic normality of the IPW estimator, we need to surrender the simplicity of the variance and give a more general result.
    
\begin{restatable}[Asymptotic normality of the IPW value estimator under misspecified models]{theorem}{vnormmis}\label{thm:4}
    In the online decision-making model with the $\varepsilon$-greedy policy, let $\V{\beta}^* = \V{\beta}_1^* - \V{\beta}_0^*$ and define
    $$
    f(\V{b}) = \int I\{\V{b}^T\V{x} \ge 0\}[\phi_1(\V{x}) - \phi_0(\V{x})] d\mathcal{P}_\V{x},
    $$
    for $\V{b}\in\mathbb{R}^d$. If Assumptions \ref{as:1}, \ref{as:2} and \ref{as:5} are satisfied, $\varepsilon_t$ is non-increasing and $\varepsilon_t\to\varepsilon_\infty>0$ as $t\to\infty$, and $f$ has continuous derivatives in the neighborhood of $\V{\beta}^*$ with $f'(\V{\beta}^*)$ bounded. Then
    \begin{equation*}
        \sqrt{t}\left(\tilde{V}_t-V^*\right)\overset{d}{\to} \mathcal{N}(0,\zeta^2),
    \end{equation*}
    where 
    \begin{align*}
        \zeta^2=&\frac{2}{2-\varepsilon_\infty}\left\{\int_{\V{\beta}^{*T} \V{x} \ge 0 }[\phi_1^2(\V{x})+\sigma_1^2] d\mathcal{P}_\V{x} + \int_{\V{\beta}^{*T} \V{x} < 0 }[\phi_0^2(\V{x})+\sigma_0^2] d\mathcal{P}_\V{x}\right\}-V^{*2}\\
        +&2f'(\V{\beta}^*)^T\Sigma^{-1}(H_0 + H_1)\Sigma^{-1}f'(\V{\beta}^*)\\
        +&\frac{4}{2-\varepsilon_\infty}\bigg[\int_{\V{\beta}^{*T} \V{x} \ge 0 }\left\{\phi_1^2(\V{x}) - \phi_1(\V{x})\V{\beta}_1^{*T}\V{x} + \sigma_1^2\right\}f'(\V{\beta}^*)^T\Sigma^{-1}\V{x} d\mathcal{P}_\V{x}\\
        &\qquad\qquad- \int_{\V{\beta}^{*T} \V{x} < 0 }\left\{\phi_0^2(\V{x}) - \phi_0(\V{x})\V{\beta}_0^{*T}\V{x} + \sigma_0^2\right\}f'(\V{\beta}^*)^T\Sigma^{-1}\V{x} d\mathcal{P}_\V{x}\bigg]
    \end{align*}
    with $H_0$, $H_1$ defined in Theorem \ref{thm:2} and $\sigma_i^2=\E(e_s^2|a_s=i)$.
    
\end{restatable}
    
The first line of the asymptotic variance is the counterpart of the asymptotic variance given in Theorem \ref{thm:3}, and the other terms are the variance introduced by the wrong decisions. 
The key to estimating $\zeta^2$ is the estimation of $f'(\V{\beta}^*)$ and we can estimate it using kernel approximation. Denote $\tilde{u}_{s,t} = y_s - \tilde{a}_s\tilde{\V{\beta}}_{1, t}^{T}\V{x}_s - (1 - \tilde{a}_s)\tilde{\V{\beta}}_{0, t}^{T}\V{x}_s$ and $\tilde{\V{\beta}}_{t} = \tilde{\V{\beta}}_{1,t} - \tilde{\V{\beta}}_{0,t}$. For each fixed vector $\V{b}$, the empirical approximation of $f(\V{b})$ is
$$
\tilde{f}_t(\V{b}) = \frac{1}{t}\sum_{s=1}^t I\{\V{b}^T\V{x}_s \ge 0\}\left\{\left(\frac{\tilde{a}_s}{\tilde{\pi}_s} - \frac{1-\tilde{a}_s}{1-\tilde{\pi}_s}\right)\tilde{u}_{s,t} + \tilde{\V{\beta}}_{t}^{T}\V{x}_s\right\},
$$
where $\{\tilde{a}_s/\tilde{\pi}_s - (1-\tilde{a}_s)/(1 - \tilde{\pi}_s)\}\tilde{u}_{s,t} + \tilde{\V{\beta}}_{t}^{T}\V{x}_s$ is a consistent estimator for $\phi_1(\V{x}_s) - \phi_0(\V{x}_s)$.
In order to estimate the derivative of $f(\V{b})$, we use the kernel smoother to smooth the empirical mean $\tilde{f}_t(\V{b})$ as
$$
\mathfrak{f}_t(\V{b}) = \frac{1}{t}\sum_{s=1}^t \Phi\left(\frac{\V{b}^T\V{x}_s}{h_t}\right)\left\{\left(\frac{\tilde{a}_s}{\tilde{\pi}_s} - \frac{1-\tilde{a}_s}{1-\tilde{\pi}_s}\right)\tilde{u}_{s,t} + \tilde{\V{\beta}}_{t}^{T}\V{x}_s\right\},
$$
where $\Phi(\cdot)$ is the cumulative distribution function of the standard normal distribution and the bandwidth $h_t \to 0$ as $t \to \infty$. Its derivative is then
$$
\mathfrak{f}'_t(\V{b}) = \frac{1}{t}\sum_{s=1}^t p\left(\frac{\V{b}^T\V{x}_s}{h_t}\right)\left\{\left(\frac{\tilde{a}_s}{\tilde{\pi}_s} - \frac{1-\tilde{a}_s}{1-\tilde{\pi}_s}\right)\tilde{u}_{s,t} + \tilde{\V{\beta}}_{t}^{T}\V{x}_s\right\}\frac{\V{x}_s}{h_t},
$$
where $p(\cdot)$ is the probability density function of the standard normal distribution. The following theorem establishes the consistency of $\mathfrak{f}'_t(\tilde{\V{\beta}}_t)$ and gives a consistent estimator for the asymptotic variance of $\tilde{V}_t$.

\begin{theorem}\label{thm:5}
Under the same conditions as in Theorem \ref{thm:4}, a consistent estimator of $f'(\V{\beta}^*)$ is $\mathfrak{f}'_t(\tilde{\V{\beta}}_t)$ with $h_t \to 0$ as $t \to \infty$. Then a consistent estimator of $\zeta^2$ is given by
\begin{align*}
    \tilde{\zeta}^2 = &\frac{2}{2-\varepsilon_t}\frac{1}{t}\sum_{s=1}^t \frac{I\{F_s\}}{P\{F_s\}}y_s^2 - \tilde{V}_t^2 + 2\mathfrak{f}'_t(\tilde{\V{\beta}}_t)^T\hat{\Sigma}_t^{-1}(\hat{H}_{0,t}+\hat{H}_{1,t})\hat{\Sigma}_t^{-1}\mathfrak{f}'_t(\tilde{\V{\beta}}_t)\\
    &+ \frac{4}{2-\varepsilon_t}\frac{1}{t}\mathfrak{f}'_t(\tilde{\V{\beta}}_t)^T\hat{\Sigma}_t^{-1}\sum_{s=1}^t\frac{I\{F_s\}}{P\{F_s\}} y_s\left\{\tilde{a}_s(y_s - \tilde{\V{\beta}}_{1, s-1}^T\V{x}_s) -(1-\tilde{a}_s)(y_s - \tilde{\V{\beta}}_{0, s-1}^T\V{x}_s)\right\}\V{x}_s,
\end{align*}
where $\hat{\Sigma}_t$, $\hat{H}_{0,t}$ and $\hat{H}_{1,t}$ are defined in Theorem \ref{thm:2} and $F_s = \{\tilde{a}_s = I\{\tilde{\V{\beta}}_{s-1}^T\V{x}_s \ge 0\}\}$.
\end{theorem}

\section{Numerical Study}\label{sec:numerical_study}
In this section, we illustrate the performance of the $\varepsilon$-greedy policies based on the online OLS and WLS estimator under both correctly specified and misspecified models.
    
\subsection{Policy Based on the Online OLS Estimator}\label{sec:numerical_c}
For the correctly specified model, we generate two covariates, so $d=3$ with the intercept. The true parameter is set to $\V{\beta}_0=(0.3,-0.1,0.7)^T$ and $\V{\beta}_1=(0.8,0.5,-0.4)^T$ for all cases. In the supplementary material, we also consider cases with higher dimensions $d = 10$ and $20$ and set half of the elements of $\V{\beta}_0$ and $\V{\beta}_1$ to zero. The goal is to show that we can do variable selection using the traditional hypothesis testing method with the help of our inferential results. The simulation results of the parameter estimators under higher dimensional cases are similar to the cases we show here. So we move them to the supplementary material due to the space limit. Each element of the i.i.d. covariates $\V{x}_t$ is sampled from a truncated normal distribution with mean zero and scale parameter one and support $[-10, 10]$ so that Assumption \ref{as:2} is satisfied.\footnote{We also experiment with discrete covariates from the discrete uniform distribution with support $\{1,2,3,4,5\}$. The results are similar so we move them to the supplementary material to avoid redundancy.} The error distributions $\mathcal{P}_{ei}$ are both the normal distribution with mean zero and variance $\sigma^2$ for $i=0,1$. We experiment with $\sigma=0.1$ and $1$ to see the influence of the scale of noise on the convergence of parameter estimator. $\varepsilon_t$ is $\{\log(t)/(10\sqrt{t})\}\wedge 1$ in our simulation. Total decision time $T=2000$ and initial random exploration time $T_0$ is $20$ for all cases. We randomly choose $10$ of the initial $20$ decisions and assign them to action 0 and others to action 1, so that both actions will have enough samples to form the initial estimators. 
    
\begin{figure}[!htbp]
    \centering
    \includegraphics[scale=.8]{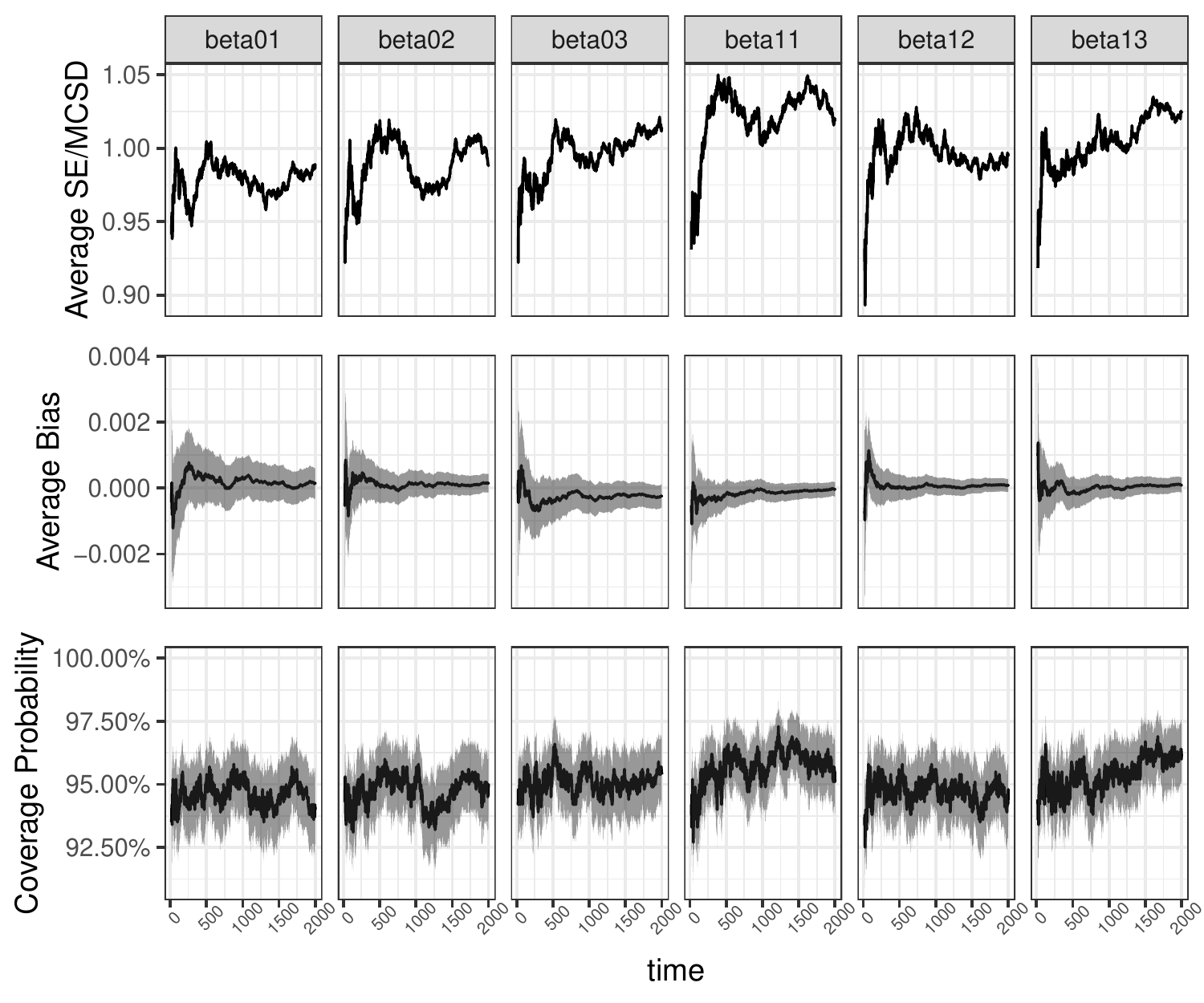}
    \caption{Convergence of the online OLS estimators with covariates from the truncated normal distribution and $\sigma=0.1$. The gray regions are 95\% confidence intervals of the Monte Carlo results.}
    \label{fig:linear_tnorm_sigma0p1_ols_parameter}
\end{figure}

\begin{figure}[!htbp]
    \centering
    \includegraphics[scale=.8]{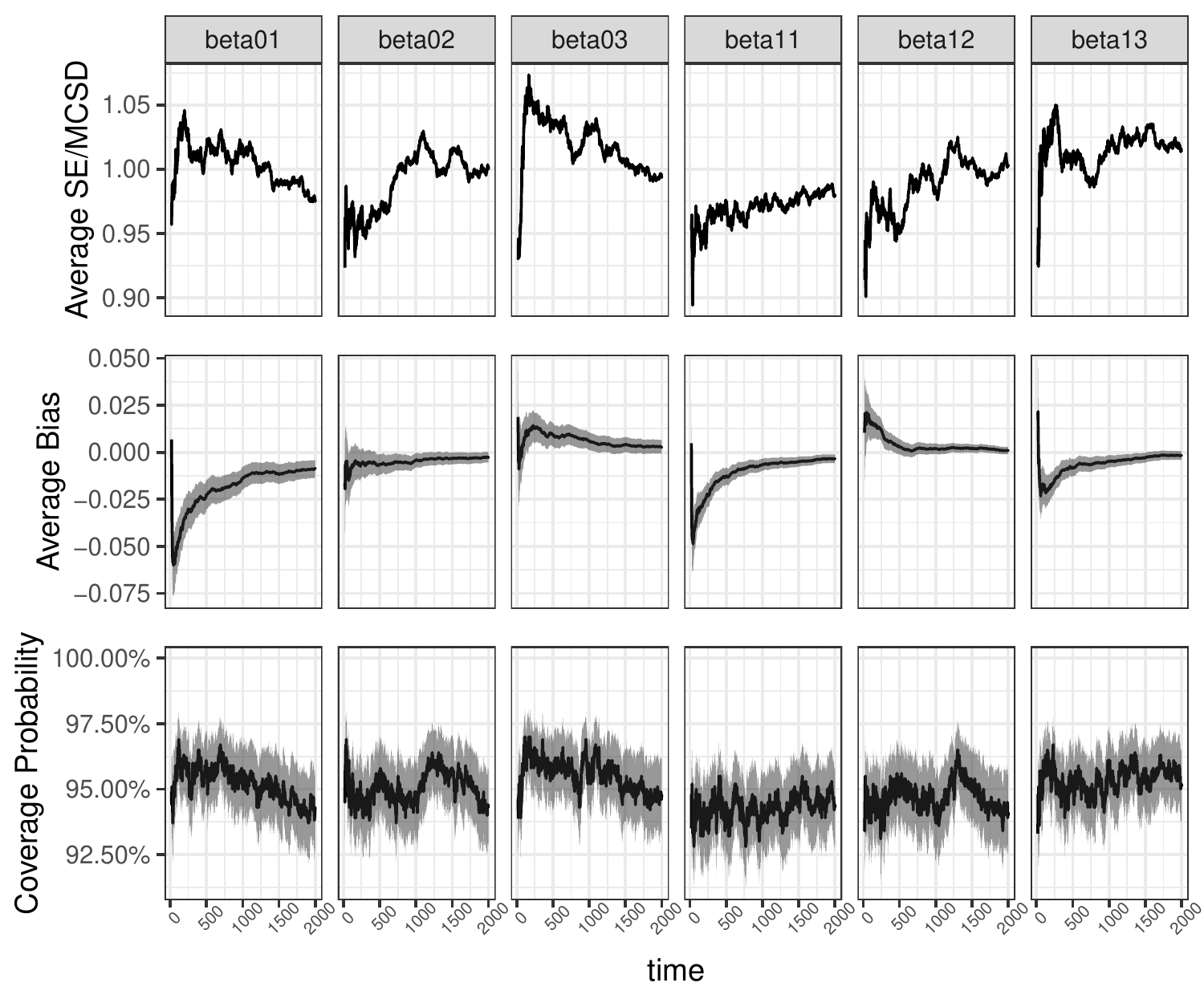}
    \caption{Convergence of the online OLS estimators with covariates from the truncated normal distribution and $\sigma=1$. The gray regions are 95\% confidence intervals of the Monte Carlo results.}
    \label{fig:linear_tnorm_sigma1_ols_parameter}
\end{figure}

We repeat the experiment 1000 times for each noise scale, and the performance of the online OLS estimators under the settings with $\sigma=0.1$ and $1$ are shown in Figures \ref{fig:linear_tnorm_sigma0p1_ols_parameter} and \ref{fig:linear_tnorm_sigma1_ols_parameter} respectively. The results reported are the ratio between the average standard error (SE) and the Monte Carlo standard deviation (MCSD) of the 1000 estimates, the estimation bias averaged over 1000 experiments and the coverage probability of the 95\% two-sided Wald confidence interval calculated by the parameter estimate and estimated standard error from each experiment.

In the case with a smaller $\sigma$ (Figure \ref{fig:linear_tnorm_sigma0p1_ols_parameter}), all of the average SE's are close to the MCSD's eventually, the average biases converge to zero, and the coverage probabilities converge to 95\%. A bigger $\sigma$ will cause bigger variation of the estimates and longer converging time. As shown in Figure \ref{fig:linear_tnorm_sigma1_ols_parameter}, the scale of the average biases are bigger than that in Figure \ref{fig:linear_tnorm_sigma0p1_ols_parameter} and the average bias of the first element of $\hat{\V{\beta}}_0$ has not converged to zero yet.
    
\begin{figure}[!htbp]
    \centering
    \includegraphics[scale=.5]{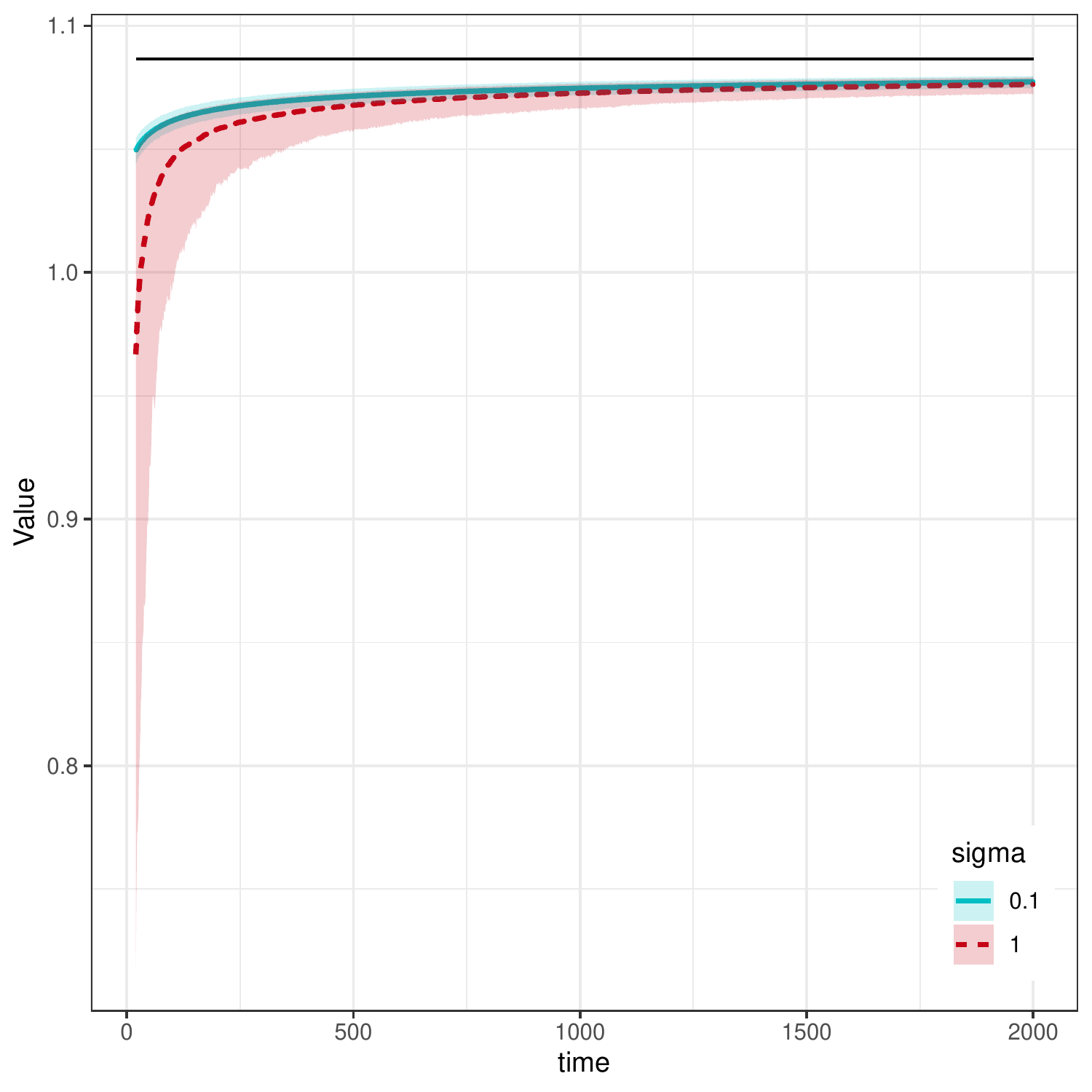}
    \caption{Convergence of expected values with covariates from the truncated normal distribution. The top straight line marks the optimal value $V$. The solid and dotted curves are expected values under the estimated policies $V_t^{\hat{\pi}}$ for $\sigma=0.1$ and $1$ respectively. 
    The bands are 2.5\% and 97.5\% quantiles of the Monte Carlo results.}
    \label{fig:linear_tnorm_sigma0p1n1_ols_evalue}
\end{figure}

For each case, we also compare the expected value functions following the oracle policy and the estimated policy, i.e, the optimal value $V$ in (\ref{eq:opt_value}) and $V^{\hat{\pi}}_t$ in (\ref{eq:est_value}). We generate $10^4$ random samples of $\V{x}$ from each case's distribution, make decisions according to the oracle and the estimated policy, and approximate the expected value by the mean of the $10^4$ expected rewards. Figure \ref{fig:linear_tnorm_sigma0p1n1_ols_evalue} shows that for the continuous covariates cases, the expected value under the estimated policy will converge to the optimal value as $t$ grows. The gap between $V$ and $V^{\hat{\pi}}_t$ is caused by the $\varepsilon$-greedy policy, which assigns about 1.7\% random actions at $t=2000$. As $t$ goes to infinity, $\varepsilon_t$ goes to zero and the gap will also disappear. Results are similar for the discrete cases. As expected, the convergence of parameter estimator and expected value function are slower for data with heavier noise $\sigma$ for all covariate distributions. 
    
\begin{figure}[!htbp]
    \centering
    \includegraphics[scale=.8]{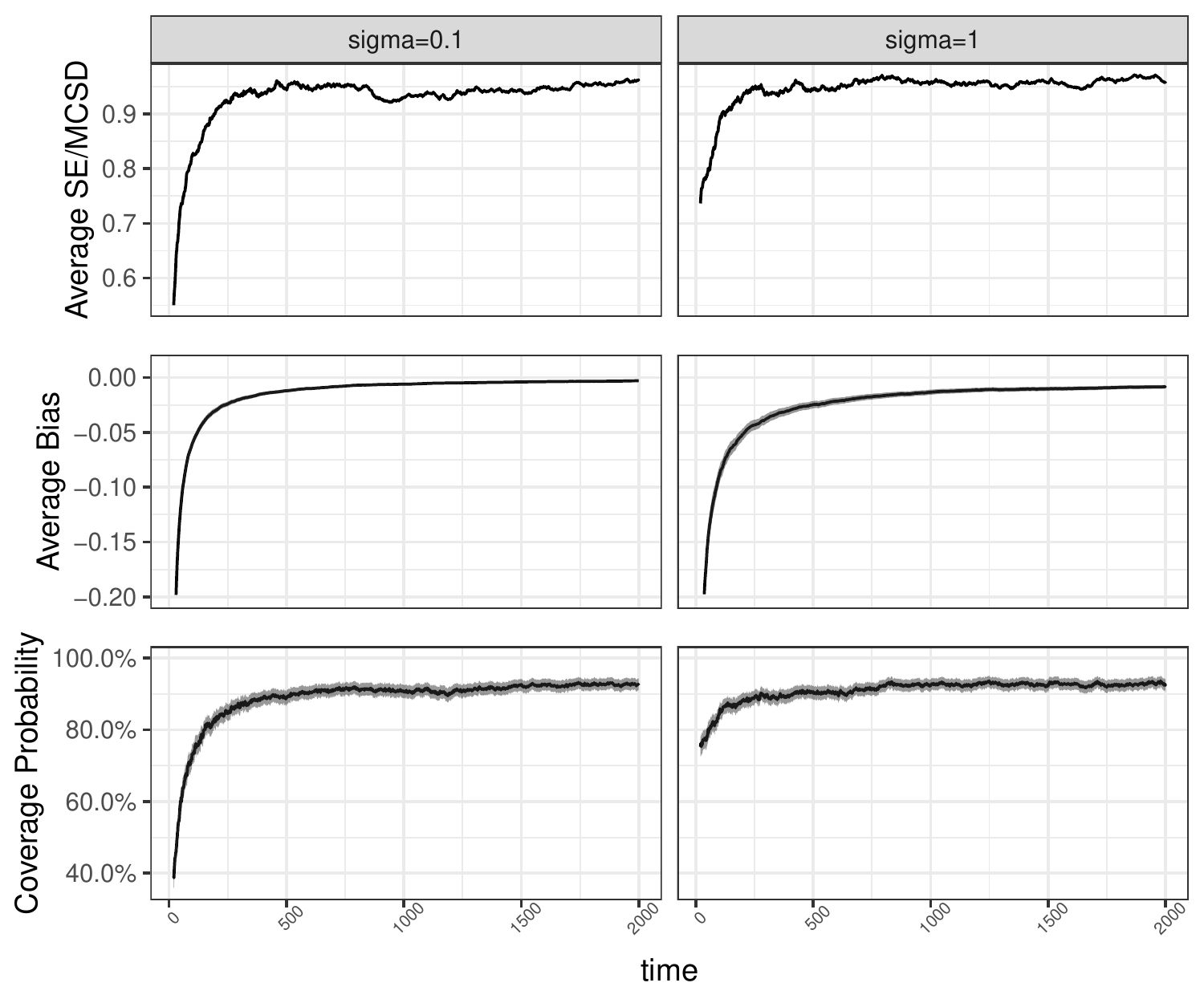}
    \caption{Convergence of the IPW estimator $\hat{V}_t$. The gray regions are 95\% confidence intervals of the Monte Carlo results.}
    \label{fig:linear_tnorm_sigma0p1n1_ols_value}
\end{figure}

The IPW estimator $\hat{V}_t$ based on the online OLS estimator is also updated online for the 1000 experiments. As shown in Figure \ref{fig:linear_tnorm_sigma0p1n1_ols_value}, its bias $\hat{V}_t - V$ converges to zero as time $t$ grows. Its average standard error is calculated according to the plug-in estimator in Theorem \ref{thm:3} and is very close to the Monte Carlo standard deviation. The coverage probability of the 95\% Wald confidence interval is around the nominal level  after 2000 steps, showing that $\hat{V}_t$ is asymptotic normal. The bias and standard error of $\hat{V}_t$ are bigger when the data noise $\sigma$ is bigger.
    
\begin{figure}[!htbp]
    \centering        
    \includegraphics[scale=.8]{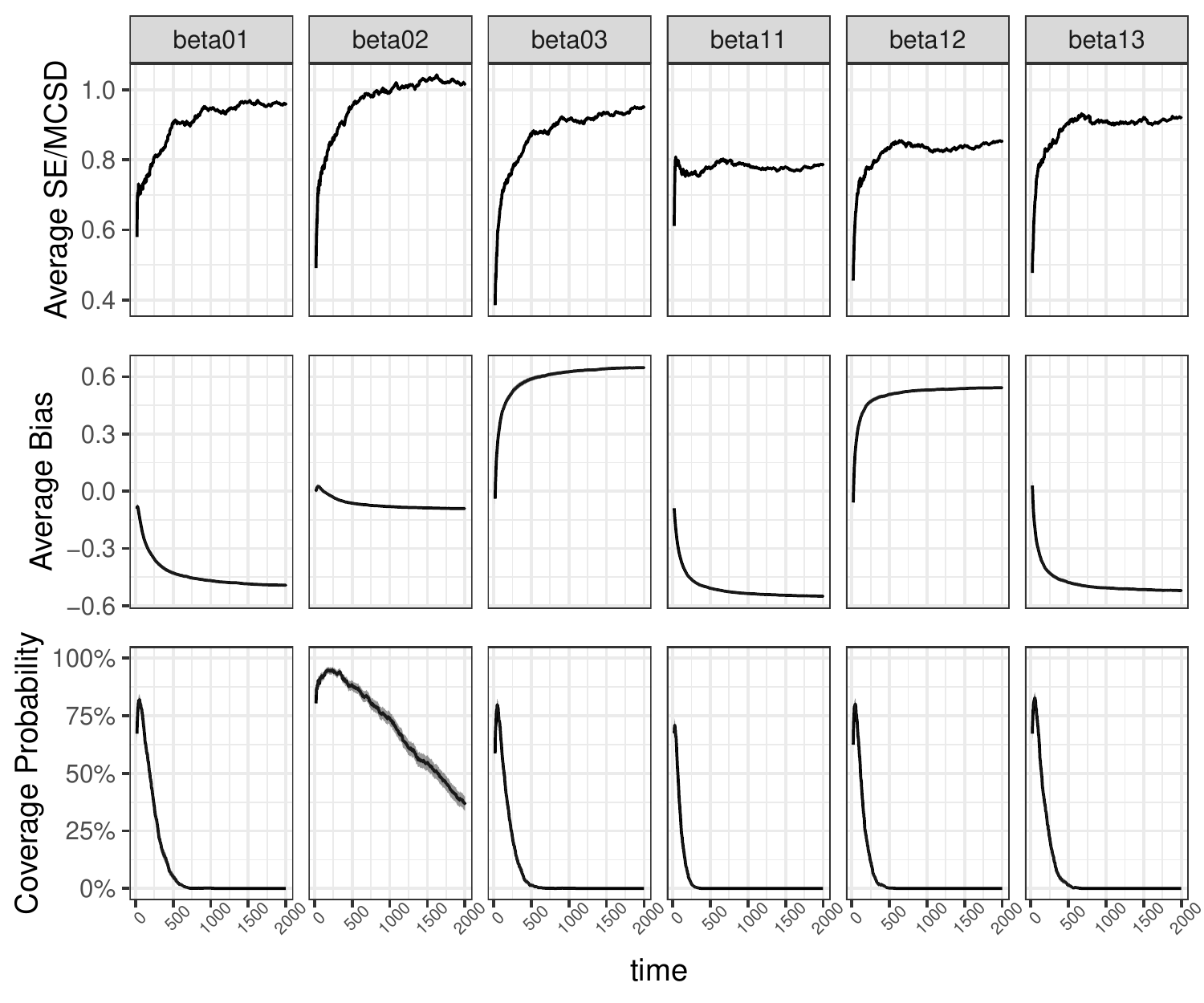}
    \caption{The performance of the online OLS estimator under misspecified model (\ref{eq:mis-model}) with $\phi_i$'s in (\ref{eq:exp}). The gray regions are 95\% confidence intervals of the Monte Carlo results.}
    \label{fig:exp_tnorm_sigma0p1_ols_parameter}
\end{figure}

We have discussed that the online OLS estimator is not unbiased when the model is misspecified. Here we illustrate it with an example. If we have true model (\ref{eq:mis-model}) with $\phi_i$'s in (\ref{eq:logis}) but want to use the online OLS estimator, then the results are shown in Figure \ref{fig:exp_tnorm_sigma0p1_ols_parameter}. Obviously, the online OLS estimator does not converge to the least false parameter, let alone the true parameter.
    
\subsection{Online WLS Estimator}\label{sec:numerical_m}
The performance of the WLS estimator are shown under both correctly specified and misspecified models. The true model for the misspecified case is set as (\ref{eq:mis-model}) with two sets of $\phi_0$ and $\phi_1$. In Setting 1 
\begin{equation}\label{eq:exp}
    \phi_i(\V{x})=\exp(\V{\beta}_i^T\V{x}), 
\end{equation}
and in Setting 2 
\begin{equation}\label{eq:logis}
    \phi_i(\V{x})=\frac{\exp(\V{\beta}_i^T\V{x})}{1+\exp(\V{\beta}_i^T\V{x})},
\end{equation}
where $\V{\beta}_0=(0.3,-0.1,0.7)^T$ and $\V{\beta}_1=(0.8,0.5,-0.4)^T$ for both sets. We consider the error term $e_t$ from $\mathcal{N}(0,\sigma)$ with $\sigma=0.1$ and $1$ for Setting 1 and $\sigma=0.1$ and $0.2$ for Setting 2 since the latter set has mean reward ranging form zero to one. The error distribution is the same for both action groups. The covariates are generated from the same truncated normal distribution as in Section \ref{sec:numerical_c} and $\varepsilon_t$ is set as a constant $\varepsilon = 0.1$.
    
\begin{figure}[!htbp]
    \centering
    \includegraphics[scale=.8]{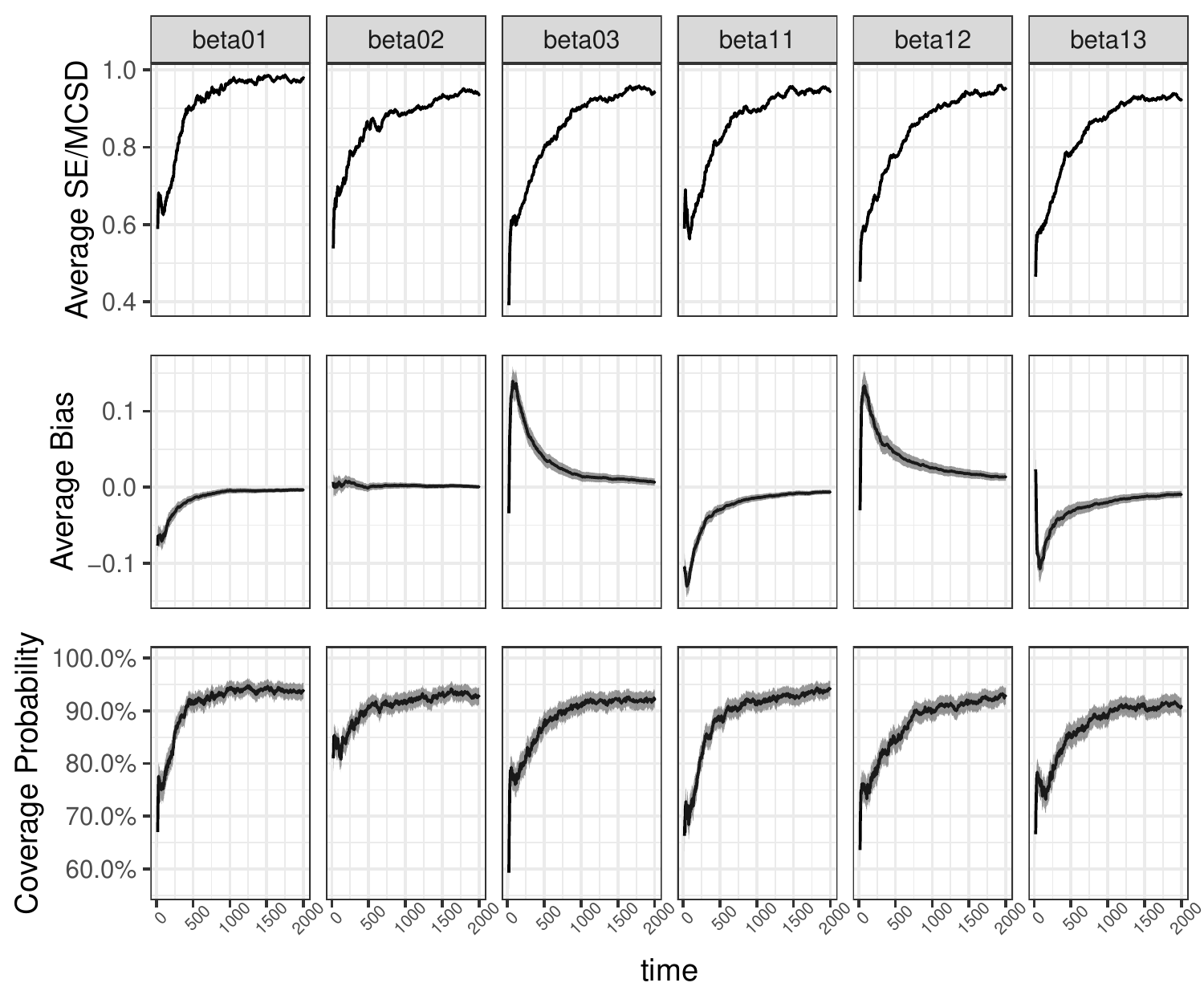}
    \caption{Convergence of the online WLS estimator under misspecified model (\ref{eq:mis-model}) with $\phi_i$'s in (\ref{eq:exp}). $\varepsilon_t=0.1$, $\sigma=0.1$. The gray regions are 95\% confidence intervals of the Monte Carlo results.}
    \label{fig:exp_tnorm_sigma0p1_wls_parameter}
\end{figure}

\begin{figure}[!htbp]
    \centering
    \includegraphics[scale=.8]{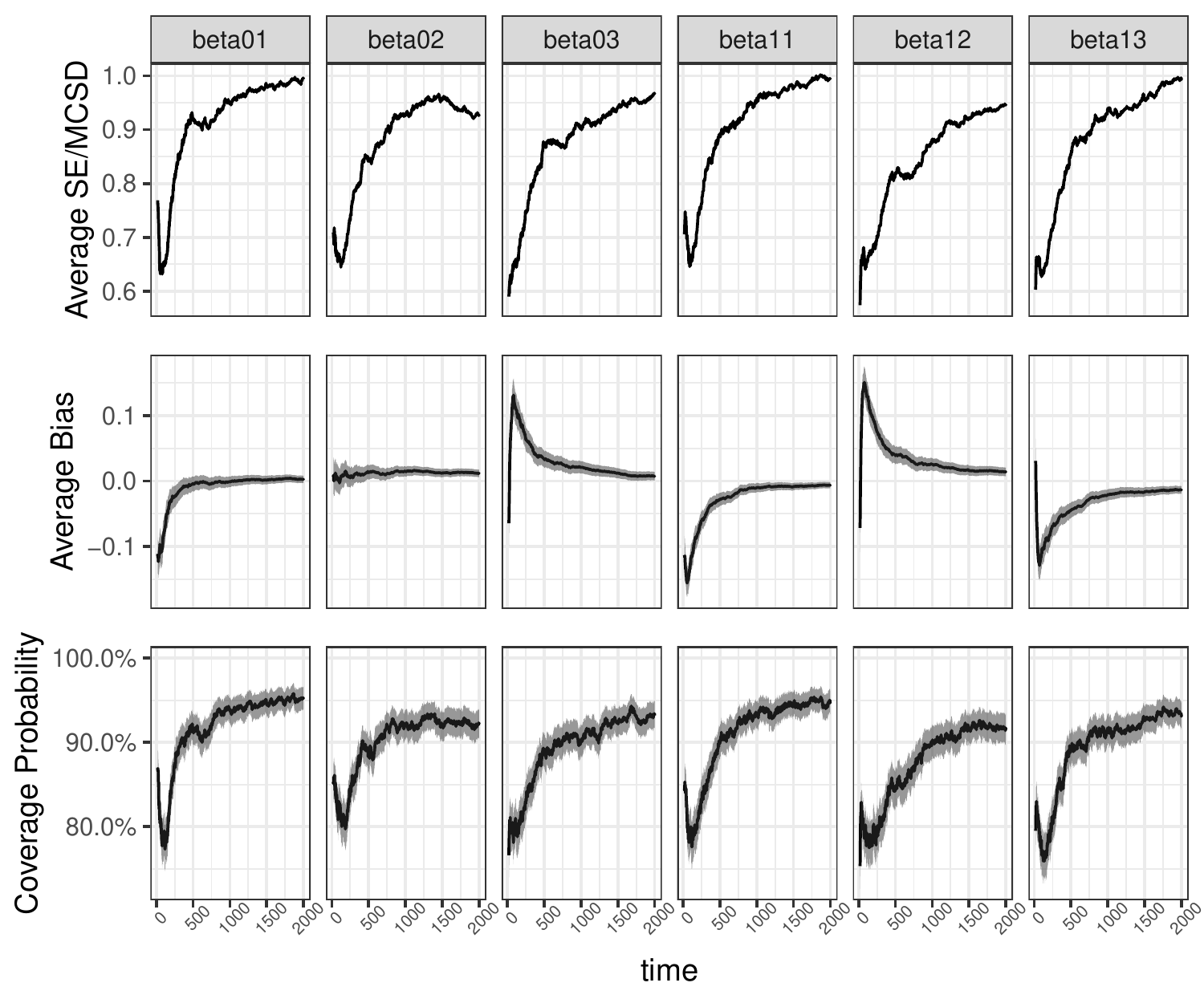}
    \caption{Convergence of the online WLS estimator under misspecified model (\ref{eq:mis-model}) with $\phi_i$'s in (\ref{eq:exp}). $\varepsilon_t=0.1$, $\sigma=1$. The gray regions are 95\% confidence intervals of the Monte Carlo results.}
    \label{fig:exp_tnorm_sigma1_wls_parameter}
\end{figure}

\begin{figure}[!htbp]
    \centering
    \includegraphics[scale=.5]{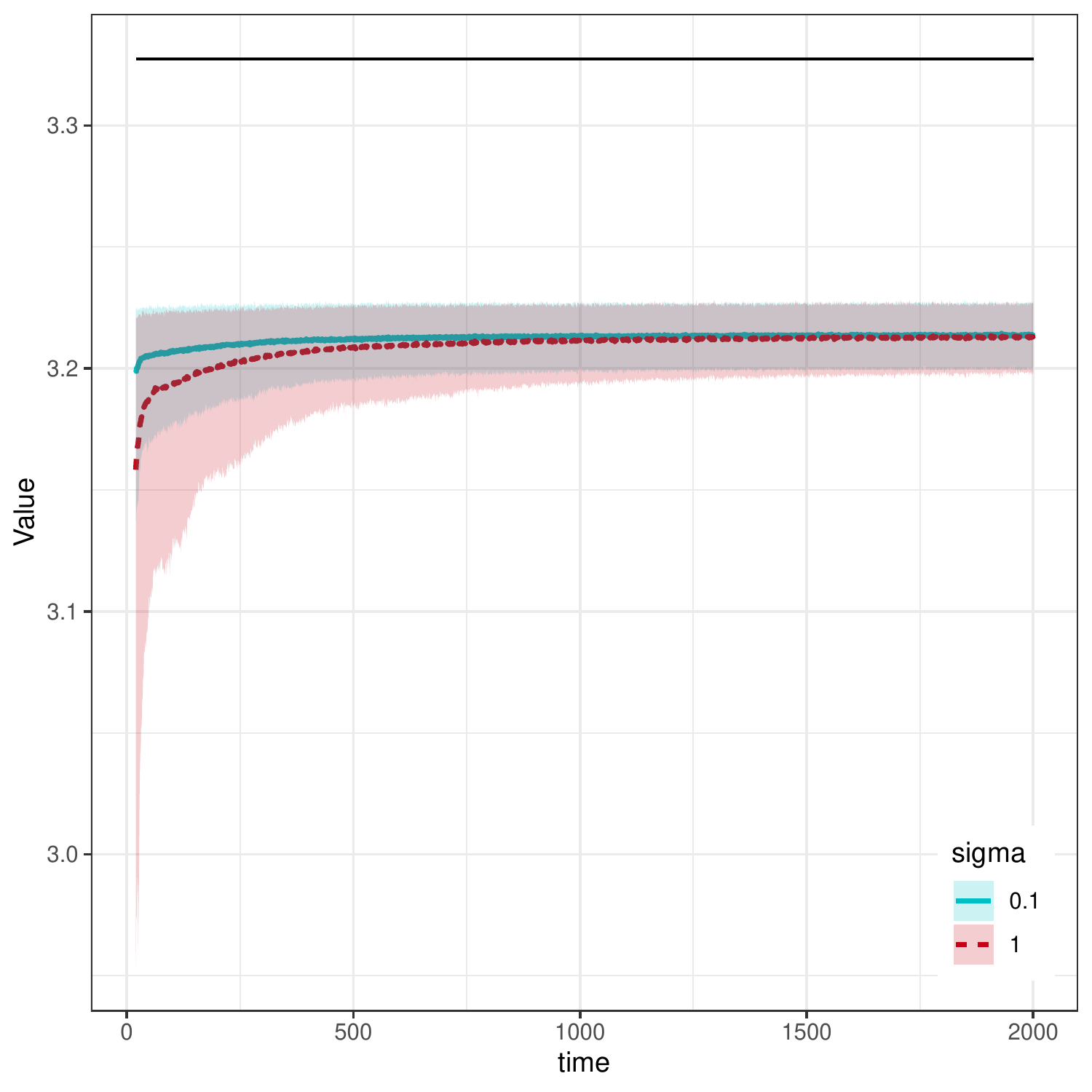}
    \caption{Convergence of expected values under misspecified model (\ref{eq:mis-model}) with $\phi_i$'s in (\ref{eq:exp}). $\varepsilon_t=0.1$. The top straight line marks the optimal value $V^*$. The solid and dotted curves are expected values under the estimated policies $V^{\tilde{\pi}}_t$ for $\sigma=0.1$ and $1$ respectively. The bands are 2.5\% and 97.5\% quantiles of the Monte Carlo results.}
    \label{fig:exp_tnorm_sigma0p1n1_wls_evalue}
\end{figure}

The experiments are also repeated 1000 times. We approximate the least false parameter by the population limit, i.e., the OLS estimator of $10^6$ random samples for each action group. Figures \ref{fig:exp_tnorm_sigma0p1_wls_parameter} and \ref{fig:exp_tnorm_sigma1_wls_parameter} demonstrate the convergence of the online WLS estimator to the least false parameter. The results are similar to that of the online OLS estimator, but it takes longer for the sandwich estimator of the variance to converge. Figure \ref{fig:exp_tnorm_sigma0p1n1_wls_evalue} shows the convergence of the expected value under the estimated policy $V^{\tilde{\pi}}_t$, which is defined in (\ref{eq:est_value_mis}), and the optimal value is now defined as $V^*$ in (\ref{eq:opt_value_mis}). Note that in the constant $\varepsilon_t$ setting, there is always a $\varepsilon/2 = 0.05$ probability that the policy will choose a non-optimal action, so the gap between $V^*$ and $V^{\tilde{\pi}}_t$ will not diminish as $t$ increases.  
    
\begin{figure}[!htbp]
    \centering
    \includegraphics[scale=.8]{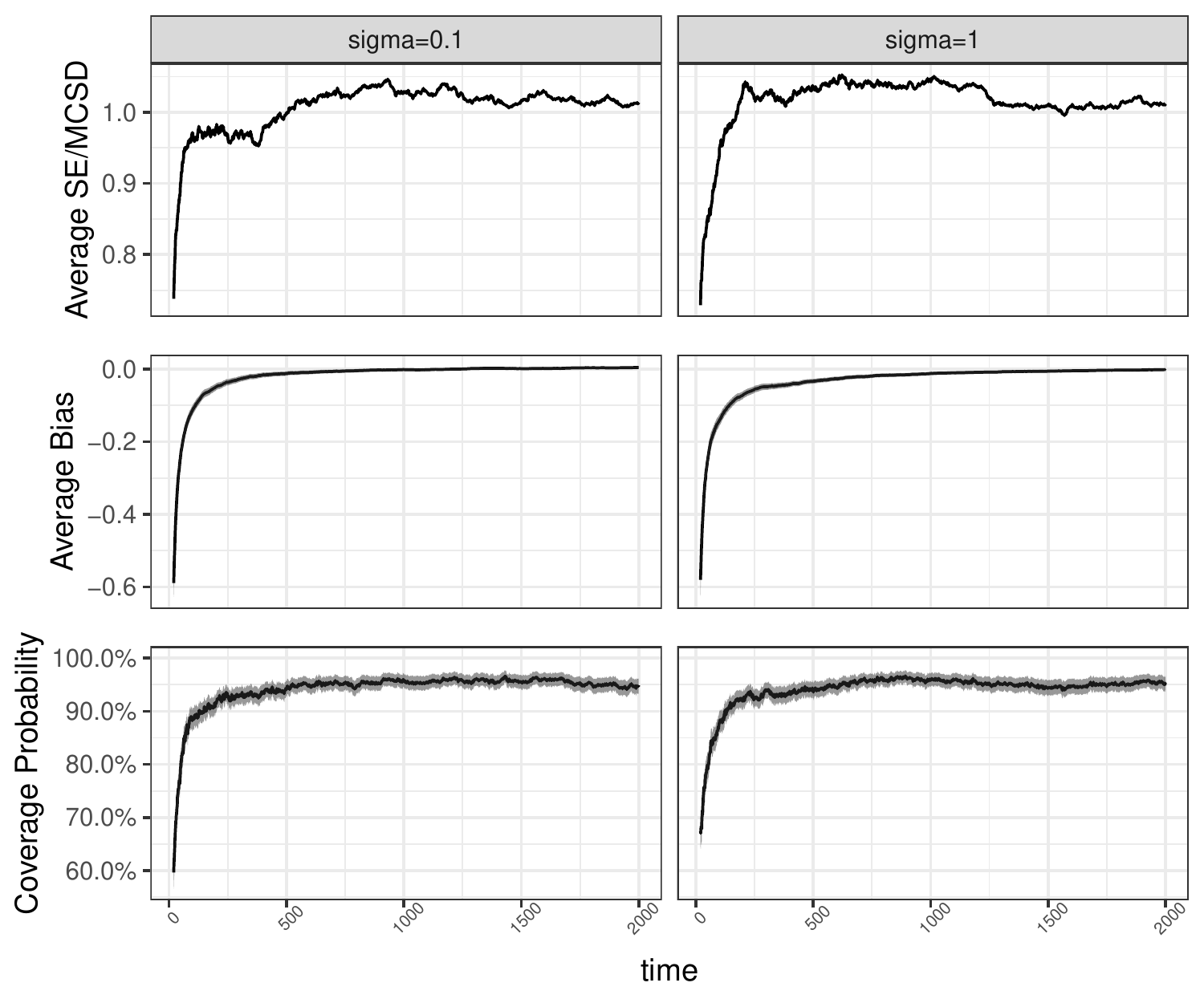}
    \caption{Convergence of the IPW estimator $\tilde{V}_t$. The gray regions are 95\% confidence intervals of the Monte Carlo results.}
    \label{fig:exp_tnorm_sigma0p1n1_wls_value}
\end{figure}

The variance of the IPW estimator $\tilde{V}_t$ is obtained by the proposed kernel estimation method. In our experiments, we take $h_t = \hat{\sigma}_t t^{-1/3}$, where $\hat{\sigma}_t$ is the sample standard deviation of $\{\tilde{\V{\beta}}_{t}^T\V{x}_s: s=1, \cdots, t\}$, and calculate the standard error of $\tilde{V}_t$ according to the estimator in Theorem \ref{thm:5}. The performance of the IPW estimator is shown in Figure \ref{fig:exp_tnorm_sigma0p1n1_wls_value}. It is shown that the bias $\hat{V}_t - V$ converges to zero as time $t$ grows and the average standard error is very close to the Monte Carlo standard deviation. The coverage probability of the 95\% Wald confidence interval is around the nominal level after 500 steps, showing that $\hat{V}_t$ is asymptotic normal. When the noise is bigger ($\sigma = 1$), the bias of $\hat{V}_t$ requires longer to diminish to zero. Only the results for model (\ref{eq:exp}) are shown here. The corresponding results for model (\ref{eq:logis}) are similar and they are shown in the supplementary material.
    
\begin{figure}[!htbp]
    \centering
    \includegraphics[scale=.8]{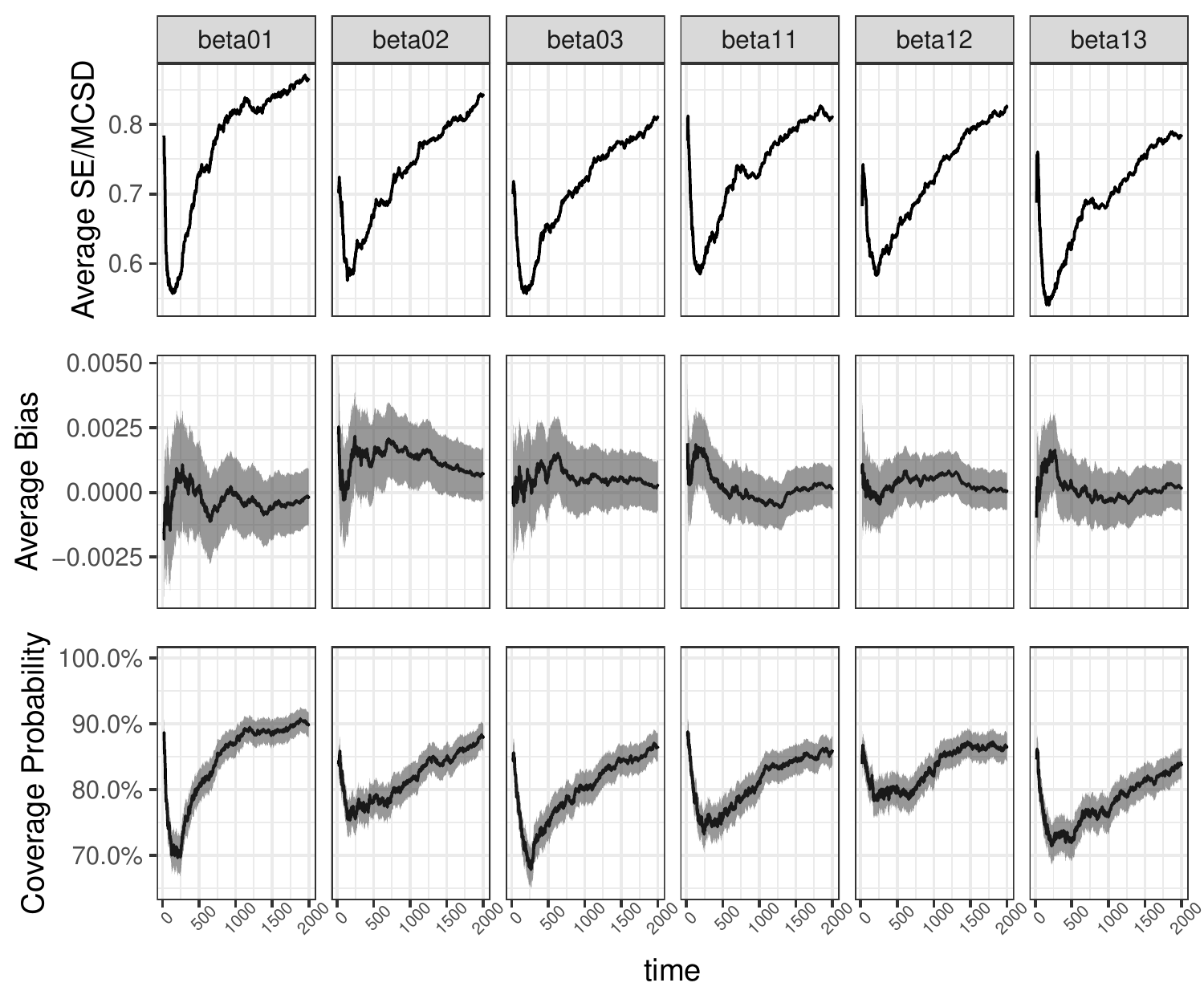}
    \caption{Convergence of the online WLS estimator under correctly specified linear model. $\varepsilon_t=(\log(t)/(10\sqrt{t})\wedge 1) \vee 0.02$, $\sigma=0.1$. The gray regions are 95\% confidence intervals of the Monte Carlo results.}
    \label{fig:linear_tnorm_sigma0p1_wls_parameter}
\end{figure}

The online WLS estimator's properties are established under a general model that includes the linear model. So it is also a candidate for estimating the parameters in a correctly specified model. We are curious to know if it is better than or as good as the online OLS estimator when the true model is linear. We experiment with the linear model setting of the previous subsection. The covariates are generated from $\mathcal{N}(0,1)$, the standard deviation of $e_t$ is set to $\sigma=0.1$, and $\varepsilon_t$ is chosen as $(\log(t)/(10\sqrt{t})\wedge 1) \vee 0.02$ so that it is bounded away from zero and comparable to the $\varepsilon_t$ studied in Section \ref{sec:numerical_c} at least for the first 1000 steps. The results for the convergence of the parameter estimators are shown in Figure \ref{fig:linear_tnorm_sigma0p1_wls_parameter}. Compared with Figure \ref{fig:linear_tnorm_sigma0p1_ols_parameter}, it can be seen that the biases of the online OLS and WLS estimators have similar scales, but the online WLS estimator is less stable due to the weighting. Its estimated standard error and hence the coverage probability converge more slowly than the online OLS estimator. As for the value convergence rate, $V^{\tilde{\pi}}_t$ based on the online WLS estimator is almost identical to $V^{\hat{\pi}}_t$ based on the online OLS estimator. This means the $\varepsilon$-greedy policies based on the two estimators give similar decisions. Therefore, if we have a good prior knowledge of the model, the online OLS estimator would be preferred as it is more efficient. Otherwise, the online WLS estimator is more robust and thus should be considered as the first choice. 
    
\section{Real Data Analysis}\label{sec:yahoo}
In this section, we used Yahoo! Today Module User Click Log Data to test the online decision-making $\varepsilon$-greedy algorithm presented above. The data set contains the news recommendation results in the first ten days of May 2009. Two mostly recommended articles on May 1st are chosen for analysis. These two articles have 405,888 recommendations in total. Here the action space are the two articles, with $a_t=1$ for recommending article ``109510'' and $a_t=0$ for recommending article ``109520''. The rewards are whether or not the readers clicked on the recommended article. We define $y_t=1$ as a click and $y_t=0$ as no click. Along with the recommendations and rewards are five continuous covariates. Each of them is between zero and one and they sum to one. So we dropped the first covariate and added a constant $x_1=1$ to form our five covariates denoted as $x_1, x_2, \cdots, x_5$. The problem of news recommendation can then be modeled as finding the decision rule that maximizes the clicking probability given the feature covariates of users.
    
With the offline data, we try to simulate the online environment by accepting only the data that match our decisions. When an entry of covariates, decision and reward is given, we apply the current estimated decision rule to the covariates and make a decision. If our decision is the same as the given decision, we will keep that entry to fit the model and update the decision rule, otherwise, we will drop it and move on to the next entry. Since the recommendations in the data set are randomized, selecting data entry like this will not introduce sampling bias.
    
\begin{figure}[!htbp]
    \centering
    \includegraphics[scale=.8]{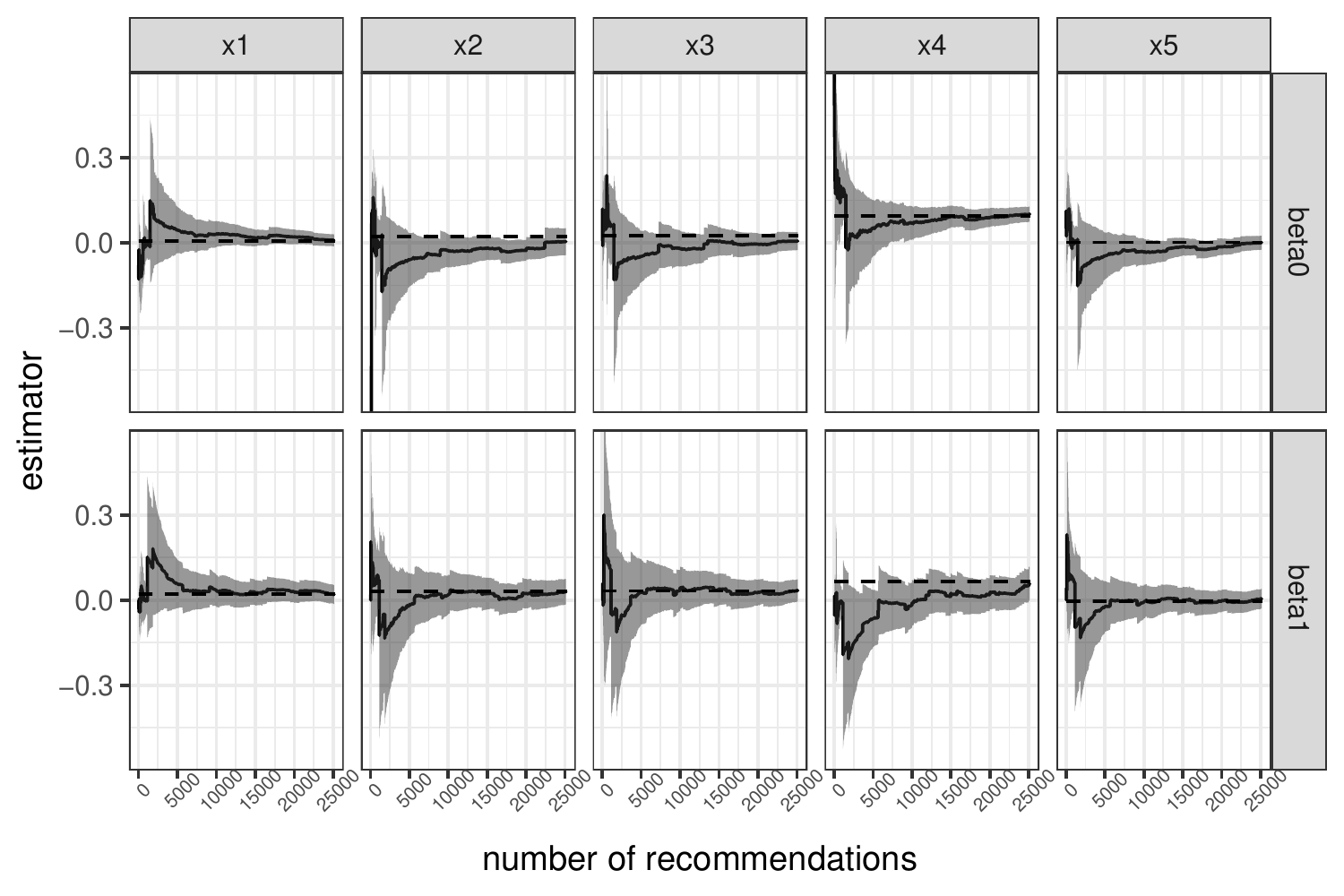}
    \caption{Convergence of the online WLS parameter estimator under the misspecified linear model. $\varepsilon_t=0.1$. $x_1=1$ and $x_2$ to $x_5$ are the four continuous covariates we kept. The dashed lines mark the least false parameters, the solid lines are WLS estimates, and the gray regions are 95\% Wald confidence intervals.}
    \label{fig:yahoo_parameter_convergence}
\end{figure}

Since the true model is unknown and most likely a non-linear model, e.g., the logistic model as in (\ref{eq:logis}), we estimate the least false parameter using the online WLS estimator proposed in the second part. The least false parameter is calculated as the population limit by fitting a linear model to all the log entries of each article. For illustration purpose, we use the first 50000 entries on May 1st to examine the performance of online decision-making, and we set $\varepsilon_t$ to be 0.1 for all $t$. Among all 50000 entries, 25085 decisions are matched. The convergence of the WLS parameter estimator is shown in Figure \ref{fig:yahoo_parameter_convergence}. In all ten parameters, only the fourth one in $\V{\beta}_0$ corresponding to $x_4$ is significantly different from 0 at 0.05 level, having an estimate of 0.101 and standard error 0.026. The third and fourth parameters of $\V{\beta}_1$ are significant at 0.1 level, all other parameters are not significant.
    
\begin{figure}[!htbp]
    \centering
    \includegraphics[scale=.5]{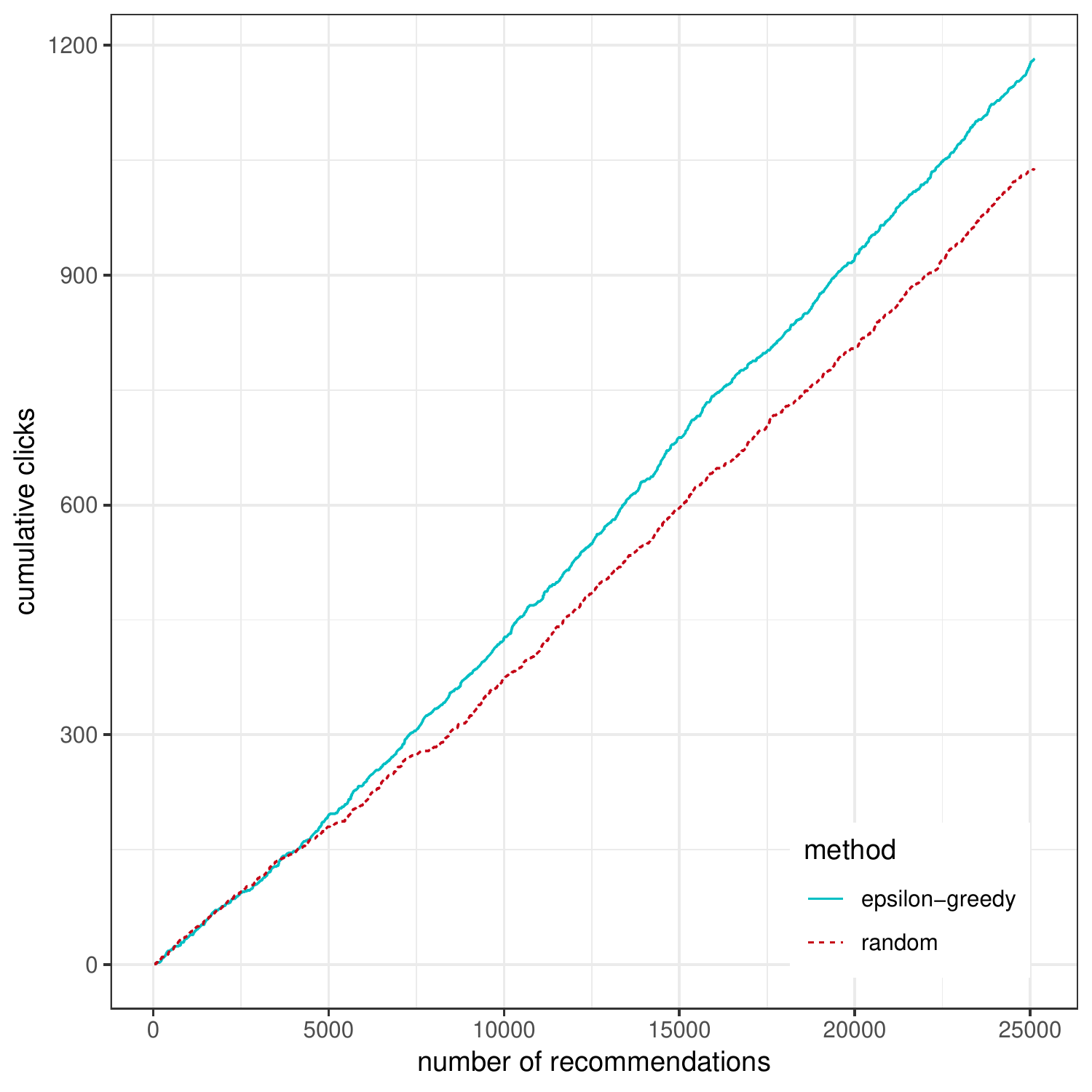}
    \caption{Cumulative performance of the $\varepsilon$-greedy policy and the random policy on the yahoo data.}
    \label{fig:yahoo_cumulative_value}
\end{figure}

In real cases, news recommendation data will be produced continually. Thus online decision-making can be utilized to learn and update the decision rule as soon as new data come in. This will save a lot of potential clicks when compared to offline learning with randomized recommendations. For the 25085 matched decisions, our $\varepsilon$-greedy policy yields a click rate of $4.71\%$, having 143 more clicks than the random policy, whose click rate is $4.14\%$ (Figure \ref{fig:yahoo_cumulative_value}). Better results can be achieved with a carefully chosen $\varepsilon_t$, and the estimated optimal click rate is $4.75\%$ using the IPW estimator.
    
\section{Discussions}\label{sec:discussions}
\subsection{Regret Bound}
In this paper we focus on the inference properties of the $\varepsilon$-greedy algorithm, but it may also be interesting to study the performance of this algorithm in terms of the regret bound, which has not been studied before. Since the regret is at least $O(t)$ when the model is misspecified, here we only consider the correctly specified linear reward setting. Specifically, we study the regret defined as the difference between the expected cumulative rewards under the oracle policy and the $\varepsilon$-greedy policy, which is
$$
R_t = \E \sum_{s=1}^t |\V{\beta}^T\V{x}_s|I\{\hat{a}_s \ne I\{\V{\beta}^T\V{x}_s \ge 0\}\},
$$
where $\V{\beta} = \V{\beta}_1 - \V{\beta}_0$. Let $\hat{\V{\beta}}_{t} = \hat{\V{\beta}}_{1, t} - \hat{\V{\beta}}_{0, t}$. The indicator function is equivalent to $|\hat{a}_s - I\{\V{\beta}^T\V{x}_s \ge 0\}|$ and can be bounded by 
$$
|\hat{a}_s - I\{\hat{\V{\beta}}_{s-1}^T\V{x}_s \ge 0\}| + |I\{\hat{\V{\beta}}_{s-1}^T\V{x}_s \ge 0\} - I\{\V{\beta}^T\V{x}_s \ge 0\}|,
$$
which separate $R_t$ into the regret from exploration 
$$
R_t^{(1)} = \E \sum_{s=1}^t |\V{\beta}^T\V{x}_s||\hat{a}_s - I\{\hat{\V{\beta}}_{s-1}^T\V{x}_s \ge 0\}|
$$
and the regret from estimation error
$$
R_t^{(2)} = \E \sum_{s=1}^t |\V{\beta}^T\V{x}_s||I\{\hat{\V{\beta}}_{s-1}^T\V{x}_s \ge 0\} - I\{\V{\beta}^T\V{x}_s \ge 0\}|.
$$
Since $\E |\hat{a}_s - I\{\hat{\V{\beta}}_{s-1}^T\V{x}_s \ge 0\}| = \varepsilon_s/2$ by (\ref{eq:pi}) and (\ref{eq:a}), $R_t^{(1)}$ is bounded by $C_1\sum_{s=1}^t \varepsilon_s$ with $C_1$ as some constant depending on $L_x$ and $d$. Under the margin assumption \ref{as:4}, we can show that $R_t^{(2)}$ is an $o_P(\sqrt{t})$ term using the same argument that we use to show (S1.3) is $o_P(1)$ in the supplementary material. Note that when we require $t\varepsilon_t^2 \to \infty$ as $t\to\infty$, there exists some constant $c>0$ such that
$$
\sum_{s=1}^t \varepsilon_s \ge \sum_{s=1}^t cs^{-\frac{1}{2}} \ge c\sum_{s=1}^t t^{-\frac{1}{2}} = c\sqrt{t}.
$$
Therefore $R_t^{(1)}$ is the dominant term and the regret bound is $O(\sum_{s=1}^t \varepsilon_s)$. In general, if we choose $\varepsilon_t = kt^{-1/2}f(t)$ with some constant $k$ and an increasing function $f(t)$ such that $f(t) > 0$ and $f(t) = o(t^{-1/2})$, we can show that 
$$
\sum_{s=1}^t \varepsilon_s \le \sum_{s=1}^t ks^{-1/2}f(t) = O(t^{1/2}f(t)).
$$
In particular, if we choose $\varepsilon_t = kt^{-\alpha}$ for some $\alpha \in (0, 1/2)$, the regret bound is $O(t^{1-\alpha})$. If $\varepsilon_t = k\log t/\sqrt{t}$, the regret bound is $O(\sqrt{t}\log{t})$. If $\varepsilon_t = k\sqrt{\log t/t}$, the regret bound is $O(\sqrt{t\log{t}})$. In comparison, the LinUCB algorithm proposed by \cite{chu2011contextual} has an regret bound of $O(\sqrt{t\log^3(t\log(t)/\delta)})$ with probability $1-\delta$, which is also achievable by our algorithm, while the force sampling method by \cite{goldenshluger2013linear} enjoys a better regret bound of $O(\log t)$.
 
From the arguments above, we can see that the decreasing rate of $\varepsilon_t$ affects both the regret bound and the tail bound of the online OLS estimator, but in different directions. If we prefer a tighter regret bound, then $\varepsilon_t$ should decrease as fast as possible. On the other hand, if we would like to estimate the reward model parameters better, a slowly decreasing or even a fixed $\varepsilon_t$ is favorable. An ideal choice of $\varepsilon_t$ should decrease as fast as possible, providing it satisfies the conditions for making inference. In practice, $\varepsilon_t = kt^{-0.499}$, $k\log t/\sqrt{t}$, or $k\log\log t/\sqrt{t}$ are all possible options when the linear reward model is correct.

\subsection{Conclusions and Future work}
The $\varepsilon$-greedy policy is a commonly adopted strategy for online decision-making problems. When compared to randomized experiments, it will greatly increase the cumulative rewards. In this work, we study the statistical inference of the model parameter and the optimal value, which is seldom discussed by antecedent literature. 

We establish the consistency of OLS estimator under the correct linear model, show its asymptotic normality and give a consistent estimator of its limiting variance. For the misspecified model, we present the WLS estimator as a consistent estimator of the least false parameter, and similar asymptotic properties are given. An important future topic would be extending our inferential results of the model parameter to other contextual bandit solutions such as Upper Confidence Bound and Thompson Sampling. Since the same data dependence structure exists in all of these exploration strategies, we can make use of the martingale structure in the online parameter estimator and study the asymptotic results.

In our setting, the optimal value, or the mean rewards under the best possible linear policy, is estimated using the IPW estimator, whose consistency and asymptotic normality have been established by us. When the model is correctly specified, we can easily estimate the asymptotic variance of the IPW estimator given the margin condition. It is also possible to construct an augmented IPW (AIPW) estimator \citep{zhang2012robust} for the optimal value and it will be more efficient than the IPW estimator. Specifically, the AIPW estimator will have the form 
$$
\hat{V}_t^{AIPW} = \frac{1}{t}\sum_{s=1}^t \left[\frac{\mathcal{I}_s}{\Pi_s}y_s - \frac{\mathcal{I}_s - \Pi_s}{\Pi_s}\{\hat{a}_s\V{x}_s^T\hat{\V{\beta}}_{1, t-1}+(1-\hat{a}_s)\V{x}_s^T\hat{\V{\beta}}_{0, t-1}\}\right],
$$
where $\mathcal{I}_s = I\{\hat{a}_s = I\{(\hat{\V{\beta}}_{1,s-1} - \hat{\V{\beta}}_{0,s-1})^T\V{x}_s \ge 0\}\}$ and $\Pi_s = \E \mathcal{I}_s$. Its asymptotic properties can be derived following similar strategies we use to show the asymptotic normality of $\hat{V}_t$ and would be interesting to study in a  future work. When the model is misspecified, the variance of the AIPW estimator can also be estimated with the kernel method.

In real cases, most of the reward models would be non-linear. So the linear model and its OLS estimator purposed by \cite{goldenshluger2013linear} will not be a suitable choice for most problems since it will cause bias under any non-randomized policy. Our online WLS estimator can fix the problem, but in a limited sense, as the decision rule can still be improved if the model is correctly specified. This demands a more flexible model, e.g., generalized linear model discussed by \citet{bastani2017exploiting}, if parametric methods are considered. Non-parametric methods \citep{yang2002randomized, qian2016kernel} are also possible solutions but it is harder to do inference. In the offline setting, non-parametric methods \citep{zhao2009reinforcement} and value search methods \citep{zhang2013robust} are proposed to address the model misspecification problem. In future work, we hope to extend these methods to the online setting to achieve better performance when the reward model is non-linear.

\bigskip
\begin{center}
{\large\bf SUPPLEMENTARY MATERIAL}
\end{center}

\begin{description}

\item[Proof of Main Results and Extended Simulation Results:] Section S1 of the supplemental material contains the proofs for all the propositions and theorems in this article. Section S2 contains the technical results and their proofs. Section S3 contains extended simulation results for discrete covariate cases, high-dimensional covariate cases and the logistic reward model setting. 

\end{description}

\bibliographystyle{apalike}
\bibliography{manuscript_rev2_unblinded}

\end{document}